\documentclass[letterpaper,12pt,titlepage,oneside,final]{book}

\newcommand{\href}[1]{#1} %

\usepackage{ifthen}
\usepackage{float}
\newboolean{PrintVersion}
\setboolean{PrintVersion}{false}

\usepackage{amsmath,amssymb,amstext} %
\usepackage[pdftex]{graphicx} %

\usepackage{booktabs}
\usepackage{graphicx}
\usepackage{colortbl}
\usepackage{multirow}
\usepackage[table]{xcolor}
\usepackage[justification=centering]{caption}
\usepackage{subcaption}
\usepackage[most]{tcolorbox}

\usepackage{soul} %
\usepackage{pifont} %

\usepackage[pdftex,pagebackref=false]{hyperref} %
\hypersetup{
    plainpages=false,       %
    unicode=false,          %
    pdftoolbar=true,        %
    pdfmenubar=true,        %
    pdffitwindow=false,     %
    pdfstartview={FitH},    %
    pdftitle={Towards Green AI: Evaluating Energy Efficiency for Quantized LLM Inference},
    pdfauthor={Petar Basta},    
    pdfsubject={Energy efficiency of quantized large language model inference},  
    pdfnewwindow=true,      %
    colorlinks=true,        %
    linkcolor=blue,         %
    citecolor=green,        %
    filecolor=magenta,      %
    urlcolor=cyan           %
}
\ifthenelse{\boolean{PrintVersion}}{   %
\hypersetup{	%
    citecolor=black,%
    filecolor=black,%
    linkcolor=black,%
    urlcolor=black}
}{} %

\usepackage[toc,abbreviations]{glossaries-extra}
\makeglossaries

\begin{document}

\pagestyle{empty}
\pagenumbering{roman}

\begin{titlepage}
        \begin{center}
        \vspace*{1.0cm}

        \Huge
        {\bf Towards Green AI: Evaluating Energy Efficiency for Quantized LLM Inference}

        \vspace*{1.0cm}

        \normalsize
        by \\

        \vspace*{1.0cm}

        \Large
        Petar Basta \\

        \vspace*{3.0cm}

        \normalsize
        A thesis \\
        presented to the University of Waterloo \\ 
        in fulfillment of the \\
        thesis requirement for the degree of \\
        Master of Mathematics \\
        in \\
        Computer Science \\

        \vspace*{2.0cm}

        Waterloo, Ontario, Canada, 2026 \\

        \vspace*{1.0cm}

        \copyright\ Petar Basta 2026 \\
        \end{center}
\end{titlepage}

\pagestyle{plain}
\setcounter{page}{2}

\cleardoublepage %
\phantomsection    %

\cleardoublepage
\phantomsection    %

 \addcontentsline{toc}{chapter}{Author's Declaration}
 \begin{center}\textbf{Author's Declaration}\end{center}

 \noindent
I hereby declare that I am the sole author of this thesis. This is a true copy of the thesis, including any required final revisions, as accepted by my examiners.
 \noindent  
  \bigskip
  
  \noindent
I understand that my thesis may be made electronically available to the public.

\cleardoublepage
\phantomsection    %

\addcontentsline{toc}{chapter}{Abstract}
\begin{center}\textbf{Abstract}\end{center}

While it is known that quantization of large language models (LLMs) reduces memory usage via lower-bit weights, studies quantifying the resulting impact on energy usage and carbon intensity are scarce. We present the first unified evaluation of weight-only quantization strategies for LLM inference across varying input/output lengths and GPUs, highlighting differences in energy efficiency in addition to accuracy degradation and runtime. We study quantized inference on Llama 2 7B, Phi 3.5 Mini Instruct, OLMo 1B, Qwen 2.5 7B, and Qwen 2.5 14B across GLUE MNLI, MMLU, HumanEval, and GSM8K datasets, evaluating 10 post-training quantization (PTQ) strategies on NVIDIA H100, A6000, A100, and L4 GPUs. We identify a novel trend showing that quantization techniques tend to exhibit peak energy efficiency relative to full-precision baselines when inputs are sufficiently long and outputs are short. Furthermore, we show that quantization strategies become marginally more energy-efficient relative to full-precision models as batch sizes increase, though gains are modest. Notably, fused-kernel implementations such as EETQ int8 and Bitsandbytes int8 offer the highest energy savings, up to 4× compared to FP32 on short text generation, with negligible accuracy loss. Finally, we observe that energy usage closely tracks runtime on our evaluated benchmarks, indicating that, in practice, latency optimization can serve as a reliable proxy for reducing the environmental footprint of LLM services. We conclude with suggested directions for strategically selecting low-carbon quantization strategies based on specific inference requirements.

\cleardoublepage
\phantomsection    %

\addcontentsline{toc}{chapter}{Acknowledgements}
\begin{center}\textbf{Acknowledgements}\end{center}

I would like to thank all the people who made this thesis possible, primarily my thesis advisor, Khuzaima Daudjee, and fellow student, Benson Guo, for their invaluable insights and assistance with this work. I would also like to thank Professors Tamer Ozsu and Sihang Liu for serving on my thesis examination committee.

\cleardoublepage
\phantomsection    %

\addcontentsline{toc}{chapter}{Dedication}
\begin{center}\textbf{Dedication}\end{center}

This thesis is dedicated to my father, Dragan Basta, for showing me that maintaining a childlike curiosity is the surest path toward a meaningful life. Also to Olga, who deserves a far greater work than this dedicated to her. Volim vas puno.

\cleardoublepage
\phantomsection    %

\renewcommand\contentsname{Table of Contents}
\tableofcontents
\cleardoublepage
\phantomsection    %

\addcontentsline{toc}{chapter}{List of Figures}
\listoffigures
\cleardoublepage
\phantomsection		%

\addcontentsline{toc}{chapter}{List of Tables}
\listoftables
\cleardoublepage
\phantomsection		%

\pagenumbering{arabic}

\chapter{Introduction}
Model quantization has emerged as a leading technique to improve memory efficiency of large language models (LLMs) \cite{quantization_memory}, significantly reducing memory footprint and traffic by representing model weights in lower precision (e.g., 8-bit integers instead of 16-bit or 32-bit floats). Building on 8-bit success, recent research has pushed to 4-bit quantization for even greater gains. Techniques like Generalized Post-Training Quantization (GPTQ) \cite{gptq} and Activation-aware Weight Quantization (AWQ) \cite{awq} claim to achieve near-lossless 4-bit compression by carefully calibrating quantization to minimize the error in each layer's outputs. The compression from quantization directly translates to lower memory usage and can unlock faster inference due to using a lower bitwidth which takes less time to transfer. 

LLMs can deliver impressive capabilities, but at high energy cost and carbon emissions \cite{deVries, carbonem1, carbonem2, carbonem3,carbon_model}. While the training phase for these models is resource-intensive, the inference phase, serving billions of queries, is projected to dominate the carbon footprint of LLMs and is already
producing 25$\times$ the carbon emissions used to train GPT-3 every single year \cite{chien}. Prior evaluations of inference using quantization strategies \cite{survey1, survey2, survey3, survey4, survey5} primarily focus on accuracy and memory efficiency, often lacking a detailed analysis or quantification of the correlating energy consumption. Some prior works report energy usage \cite{survey6, survey7, survey8}, but their scope is constrained, with narrow dataset coverage and limited analysis of performance across differing workloads. In contrast, our work systematically benchmarks energy usage across multiple quantization methods, offering the first comprehensive evaluation that directly connects quantization choices with energy efficiency outcomes, including contributions from input and output stages.

The energy efficiency benefits of quantization stem from multiple factors. First, a smaller model (in memory) means less data to transfer during each step from GPU memory to compute units, which improves utilization and mitigates memory-bound stalls \cite{pytorch}. Second, specialized low-bit kernels can compute matrix multiplications using int4/int8 arithmetic, which processes more operations per cycle (e.g., via NVIDIA Tensor Cores that support 4-bit and 8-bit dot products), thus finishing computations faster \cite{int4}. Faster execution often consumes less energy per task since the GPU spends less time active, even if instantaneous power draw may be similar. However, achieving these gains in practice depends on the implementation. If the quantized inference pipeline introduces high overhead (e.g., per-token dequantization costs, or inefficient kernel launches), the theoretical savings may not materialize. Recent developments have looked into even quantizing the KV cache to reduce memory and energy, with 4-bit KV entries yielding up to $2–3\times$ less memory use for long contexts \cite{hf}. In this work, we focus on weight quantization, but we note that comprehensive solutions to LLM efficiency may combine weight, activation, and cache quantization. Further, we exclusively analyze PTQ methods as testing quantization-aware training (QAT) methods would require retraining large models from scratch, which is computationally infeasible for our setup.

In sum, energy-efficient LLM inference requires balancing compute, accuracy, and memory considerations, with quantization offering a promising path to cut both memory usage and compute time. The key questions we investigate are: How do different quantization strategies compare in energy per token across tasks? Which methods best handle short outputs versus long outputs? And how do kernel fusion and configuration details impact real-world energy savings? By evaluating a range of 4-bit and 8-bit approaches on multiple models and tasks, we aim to provide actionable insights for deploying LLMs with lower carbon footprints.

\section{Thesis Contributions}
This thesis advances the literature on LLM inference optimization by providing the first holistic, empirical comparison of weight-only post-training quantization strategies with a primary focus on energy efficiency. While prior work has largely emphasized memory reduction and accuracy impacts, this study systematically measures energy consumption, runtime, and accuracy across varying input/output lengths, batch sizes, model architectures, and GPU platforms. Through this analysis we identify several key findings for practitioners to refer to when implementing or improving real-world systems.

\section{Thesis Organization}
The remainder of this thesis is organized as follows:

Chapter 2 establishes the theoretical foundations necessary to understand the scope of this work. It reviews the inference pipeline of large language models, details the Transformer architecture, and provides a technical overview of quantization principles, including symmetry and granularity.

Chapter 3 surveys the related work in the field, discussing existing studies on model compression and energy analysis. It identifies gaps in the current literature, specifically the lack of evaluations linking quantization to energy consumption and positions this thesis within the broader research landscape.

Chapter 4 presents a detailed technical analysis of the seven post-training quantization techniques evaluated in this study: Quanto, HQQ, TorchAO, Bitsandbytes, AutoAWQ, GPTQ, and EETQ. For each strategy, the chapter examines the underlying algorithmic approach, the specific mechanisms used for inference, and the resulting trade-offs between computational overhead and model accuracy.

Chapter 5 contains the comprehensive empirical evaluation. It details the experimental methodology and presents results regarding energy efficiency, runtime speedups, accuracy degradation, and memory usage across various models, datasets, and GPU architectures. It also analyzes the correlation between runtime and energy usage and estimates the cost and carbon emissions associated with different strategies.

Chapter 6 concludes the thesis by summarizing the key findings and their implications for environmentally aware AI. It also outlines potential avenues for future work, such as context-aware scheduling and extending quantization beyond model weights.

\chapter{Background}

This chapter establishes the theoretical foundations necessary to analyze the efficiency of large language models (LLMs). We first review the inference stage of LLMs, followed by the Transformer architecture that underpins these models, and finally a detailed overview of the quantization principles used to reduce their memory footprint and computational cost.

\section{LLMs for Inference}

Deploying LLMs involves a multi-stage pipeline that transforms input text into generated output. Unlike the training phase, which focuses on weight updates and typically maximizes throughput via large batches, inference is often latency-sensitive and operates in two distinct phases: prefill and decode.

The prefill phase occurs when the model receives an input prompt. During this stage, the model processes all input tokens in parallel. Because the entire prompt is available at once, the underlying GPU operations consist of large matrix-matrix multiplications. 

Following prefill, the model enters the decode phase which generates output tokens. Since LLMs are autoregressive, they generate output one token at a time. The model predicts the next token, appends it to the sequence, and feeds it back as input for the subsequent step. Unlike prefill, this phase involves matrix-vector multiplications because each decoding step processes a single newly generated token, resulting in a token representation that is multiplied against the model's weight matrices.

\section{Transformer Architecture}

The Transformer architecture was introduced by Vaswani et al. for sequence modeling tasks \cite{attention}. It is based solely on attention mechanisms and feed-forward networks. The Transformer processes an input sequence by first embedding the tokens and adding positional encodings to retain word order. Each Transformer layer consists of a multi-head self-attention mechanism followed by a feed-forward network, with residual connections and layer normalization applied around each sublayer. The encoder stack maps the entire input sequence to a sequence of contextual representations that capture relationships among all tokens.

Transformers are widely used in large language models because they enable parallel computation and capture long-range dependencies. In a Transformer, self-attention connects every pair of positions in the input with a constant number of sequential operations, in contrast to recurrent networks whose computation steps scale with sequence length. This design is more parallelizable and trains faster than recurrent architectures. As a result, very large Transformer-based models can be trained efficiently on vast datasets, which attain state-of-the-art performance on many language tasks.

\subsection{Attention Mechanisms}

Attention mechanisms allow the model to weigh the importance of different tokens. In the Transformer, an attention function maps queries and keys to an output via a weighted sum of values. Specifically, the Transformer uses scaled dot-product attention. Given a query matrix $Q$, a keys matrix $K$, and a values matrix $V$, the attention output is described as 
\[
\mathrm{Attention}(Q,K,V) = \mathrm{softmax}\Bigl(\frac{QK^\top}{\sqrt{d_k}}\Bigr)\,V
\] 
where $d_k$ is the dimension of the keys. The softmax produces weights that reflect the relevance of each key to a query. In self-attention, the queries, keys, and values all come from the same input sequence, allowing each position to aggregate information from all positions. This enables the model to capture contextual relationships directly, regardless of distance in the sequence.

The Transformer further enhances its representational power through multi-head attention, where each head captures different features of the input. By processing this information in parallel and then concatenating the results, the model can simultaneously attend to multiple relationships within the input context, improving its ability to model complex patterns in language.

\subsection{Encoder vs. Decoder}

The Transformer architecture includes an encoder and a decoder. The encoder consists of a series of identical layers; each encoder layer has two sublayers: multi-head self-attention and a feed-forward network. The encoder reads the entire input sequence in parallel and produces a set of contextualized embeddings that capture relationships between tokens. Encoder-only models (such as BERT \cite{bert}) use only this stack of layers without a decoder.

The decoder is also a stack of identical layers, but each layer has three sublayers. Each decoder layer is made up of a masked multi-head self-attention layer, a multi-head attention over the encoder's output, and a feed-forward network. The masked self-attention in the decoder ensures that each position can only attend to earlier positions, enforcing auto-regressive generation. Decoder-only models (such as GPT \cite{gpt}) omit the separate encoder and cross-attention sublayer and apply masked self-attention to the input prompt. During inference with a decoder model, there is a prefill phase, processing the input context in parallel, and a decode phase, generating tokens one by one. In practice, models first process the input prompt (the prefill phase) to initialize its key–value memory, and then generate each new token sequentially, conditional on all previously generated tokens.

\section{Quantization Background}
Quantization is a technique used to reduce the memory footprint and computational cost of LLMs by mapping floating-point weights (e.g., \texttt{float32}) to lower precision representations (e.g., \texttt{int8}). During quantization, a real-valued input $x$ is linearly mapped to a discrete integer representation using a scale factor $S$ and an optional zero-point $Z$. The quantization parameters are typically defined as:

\begin{equation}
S = \frac{v_{\max} - v_{\min}}{q_{\max} - q_{\min}}, \qquad
Z = q_{\min} - \frac{v_{\min}}{S}
\end{equation}

so that $x = v_{\min}$ maps to $q_{\min}$ and $x = v_{\max}$ maps to $q_{\max}$.

\subsection{Symmetric vs. Asymmetric Quantization}

In symmetric quantization, the value range is centered around zero, i.e., $v_{\min} = -v_{\max}$. In this case, the zero-point is set to zero ($Z = 0$), and only a single scale parameter $S$ is needed to map the original values to $[-q_{\max},q_{\max}]$:

\begin{equation}
x_q = \mathrm{round}\left(\frac{x}{S}\right)
\end{equation}

In asymmetric quantization (also called affine quantization), the input range $[v_{\min},v_{\max}]$ does not have to be centered at zero. The zero-point $Z \neq 0$ is chosen (e.g., via calibration) such that real 0 is accurately mapped to an integer offset $Z$:

\begin{equation}
x_q = \mathrm{round}\left(\frac{x}{S}\right) + Z
\end{equation}

This distinction is important because symmetric quantization simplifies hardware implementation and can result in better runtime efficiency, while asymmetric quantization can more accurately represent data that is primarily positive or not centered cleanly around zero. In practice, all the quantization strategies we assess in this paper use symmetric quantization by default.

\subsection{Quantization Granularity}

Quantization granularity refers to the scope over which a set of quantization parameters (scale and zero-point) is shared. Different granularities trade off memory and compute efficiency for accuracy.

\noindent \textbf{Per-Tensor Quantization}

Per-tensor quantization uses a single scale and zero-point for an entire tensor (e.g., all weights of a layer). This is the simplest variation since every value in the tensor is quantized uniformly using the same mapping. While it minimizes metadata and is easy to implement, it may suffer from substantial accuracy degradation if values across different channels vary widely in magnitude.

\noindent \textbf{Per-Channel Quantization}

Per-channel quantization assigns separate quantization parameters to each output channel (row/column of the weight matrix, depending on convention). In convolutional and linear layers, this typically means each output neuron is quantized with its own scale and zero-point. This approach better preserves the relative precision of weights across channels, particularly when their distributions differ significantly. The main trade-off with this approach is increased memory and compute overhead, as additional scale and zero-point values must be stored and calculated per channel.

\noindent \textbf{Per-Group (Block) Quantization}

Per-group, or block-wise, quantization generalizes per-channel quantization by grouping weight elements (or channels) into fixed-size blocks, each sharing a common scale and zero-point. Typical group sizes tend to be in the range of 32 to 128. Smaller group sizes improve accuracy by using tighter mappings to local variations in the weights, whereas larger group sizes can lower memory usage and computational costs.

\chapter{Related Work}

This chapter is intended to provide context of existing surveys on different quantization strategies and their energy usage.

Li \textit{et al.} (2024) perform one of the first large-scale evaluations of post-training quantization on LLMs~\cite{li2024evaluating}. They apply weight, activation, and KV-cache quantization to 11 model families (125M--180B parameters) across diverse tasks (basic NLP, emergent abilities, trust, dialogue, long-context). Their key findings include that moderate quantization (e.g., 4-bit weights, 8-bit activations, 4-bit KV) generally causes negligible accuracy loss. They also observe that larger models tolerate more aggressive weight/KV quantization (so a big model at 3-bit can outperform a smaller model at higher precision under the same memory)~\cite{li2024evaluating}. However, Li \textit{et al.} focus primarily on accuracy across benchmarks and do not evaluate real-world inference efficiency, GPU latency, or energy measurements, nor do they examine variability across different hardware or sequence lengths. In contrast, this thesis extends their evaluation by systematically benchmarking quantized LLM inference on multiple different GPUs, including latency and energy consumption as well as accuracy. This makes our study the first to jointly consider accuracy, latency, and energy tradeoffs in quantized LLMs across multiple hardware and model configurations.

Jin \textit{et al.} (2024) propose a structured framework for evaluating quantization in LLMs along three axes (knowledge/capacity, alignment, and efficiency)~\cite{survey1}. They conduct experiments on ten diverse benchmarks and find that 4-bit quantization typically retains performance comparable to the full-precision models. They also show that larger-parameter quantized LLMs generally outperform smaller ones, reinforcing the notion that model scale can offset quantization loss~\cite{survey1}. Importantly, Jin \textit{et al.} note that quantization yields memory savings but often slows down inference (i.e., a speed/memory tradeoff). While they highlight this efficiency issue, they do not provide detailed hardware-level measurements of inference time or power draw. In particular, they do not quantify energy usage or study performance variability across different GPUs or sequence lengths. Our work builds on these insights by explicitly measuring the hardware efficiency of quantized inference across different models and GPUs. This allows us to fill the gap left by Jin \textit{et al.}, namely, to report how various quantization schemes trade off accuracy and energy/latency in practice.

Hu \textit{et al.} (2024) perform a cluster-level analysis of six months of LLM development workloads in a production GPU datacenter~\cite{hu2024characterization}. They characterize how LLM training and evaluation jobs differ from traditional deep learning, finding that many LLM jobs are relatively short evaluation tasks (leading to queuing delays) and that resource usage is highly imbalanced (e.g., pretraining jobs use the vast majority of GPU-hours while evaluation jobs dominate job count)~\cite{hu2024characterization}. Crucially, they observe that GPUs dominate power consumption in the cluster, roughly two-thirds of server power is drawn by GPUs (with CPUs and PSUs consuming $\sim$11.2\% and 9.6\% respectively)~\cite{hu2024characterization}. They use these observations to motivate fault-tolerant training and smarter scheduling for LLMs. However, Hu \textit{et al.}'s study is at the systems/datacenter level and does not examine model quantization or inference at the user level. They do not investigate how quantization affects per-query latency or energy on different GPU types. Our benchmark complements their work by drilling down to the model-inference level: we measure how quantizing LLMs impacts both performance and energy efficiency on modern GPUs. We bridge the high-level datacenter perspective of Hu \textit{et al.} with fine-grained inference evaluation, establishing the first benchmark for quantized LLMs that explicitly measures energy consumption alongside accuracy and latency.

Alizadeh \textit{et al.} (2024) provide a broad experimental analysis of energy consumption and accuracy trade-offs specifically for software development tasks~\cite{survey8}. They evaluate model families across four distinct coding activities: code generation, bug fixing, docstring generation, and test generation. Their findings indicate that energy usage varies significantly by task type and that larger models often yield diminishing returns in accuracy relative to their power costs. However, Alizadeh \textit{et al.} treat quantization as a uniform variable, relying exclusively on the GGUF format used by the Ollama runtime without distinguishing between the underlying algorithms. They do not investigate how different post-training quantization techniques or their specific kernel implementations influence efficiency. In contrast, our work shifts the focus from model selection to strategy selection: we systematically benchmark ten distinct quantization methods and analyze how specific algorithmic choices and input/output token ratios drive energy efficiency, providing the granular technical analysis of quantization overhead that Alizadeh \textit{et al.} overlook.

\chapter{Quantization Techniques}

This chapter provides a detailed technical analysis of the seven PTQ techniques selected for evaluation in this study. These methods represent a diverse cross-section of the current state-of-the-art, varying in their optimization objectives (weight reconstruction vs. output error minimization), target bit-widths (4-bit vs. 8-bit), and backend implementations (native PyTorch operations vs. custom fused kernels). For each strategy — Quanto, HQQ, TorchAO, Bitsandbytes, AutoAWQ, GPTQ, and EETQ — we examine the underlying algorithmic approach, the specific mechanism used for inference, and the resulting trade-offs between computational overhead and model accuracy.

We also provide a breakdown of supported symmetry options by quantization strategy in Table~\ref{tab:quant_symmetry} as well as supported granularities in Table~\ref{tab:quant_granularity}.

\begin{table}[H]
\centering
\begin{tabular}{|l|c|c|}
\hline
\textbf{Quantization Technique} & \textbf{Symmetric} & \textbf{Asymmetric} \\
\hline
Quanto         & \ding{51} & \ding{51} \\
HQQ            & \ding{51} & \ding{51} \\
TorchAO        & \ding{51} & \ding{51} \\
Bitsandbytes   & \ding{51} & \ding{55} \\
EETQ           & \ding{51} & \ding{55} \\
AutoAWQ        & \ding{51} & \ding{51} \\
GPTQ           & \ding{51} & \ding{51} \\
\hline
\end{tabular}
\caption{Symmetric vs.\ Asymmetric Support Across Quantization Techniques}
\label{tab:quant_symmetry}
\end{table}

\begin{table}[H]
\centering
\begin{tabular}{|l|c|c|c|}
\hline
\textbf{Quantization Technique} & \textbf{Per-Tensor} & \textbf{Per-Channel} & \textbf{Per-Group} \\
\hline
Quanto & \ding{51} & \ding{51} & \ding{51} \\
HQQ & \ding{55} & \ding{51} & \ding{51} \\
TorchAO  & \ding{51} & \ding{51} & \ding{51} \\
Bitsandbytes & \ding{55} & \ding{51} & \ding{51} \\
EETQ & \ding{55} & \ding{51} & \ding{55} \\
AutoAWQ & \ding{55} & \ding{55} & \ding{51} \\
GPTQ & \ding{55} & \ding{51}& \ding{51} \\
\hline
\end{tabular}
\caption{Granularity Support Across Quantization Techniques}
\label{tab:quant_granularity}
\end{table}

\section{Quanto}
Quanto \cite{quanto} is a PyTorch-based quantization strategy designed for flexibility and ease of use. 

\subsection{Quantization Approach}
Quanto implements linear weight quantization in multiple precisions (int8, int4, int2, and float8). Quanto quantization can be applied per-tensor or per-channel, with per-channel default for int8 and float8, to better capture each layer's weight distribution, per-group quantization as a default for lower bitwidths such as int4. By default, weight quantization is symmetric (zero-point 0), using simple statistics like min/max or absolute max as the scaling factor. For example, 4-bit weights are linearly mapped into 16 levels (-8 to 7) by dividing each weight's range by a scale and rounding. 

Quanto supports very low bit-widths (2-4 bits) by allowing group-wise quantization: weights are divided into groups that each get their own scale/zero-point to reduce quantization error. Overall, the initial PTQ procedure in Quanto closely follows PyTorch's standard approach: insert quantize/dequantize hooks around layers and determine scales/zero-points via calibration or known distributions.

\subsection{Quantized Inference Mechanism}
Once calibrated, Quanto automatically inserts quantized modules (e.g., QLinear, QConv2d) to replace float operations. Each module stores quantized weights along with the per-tensor, per-group or per-channel scales (and zero-points if asymmetric) needed to reconstruct them. On CUDA devices, Quanto uses low-level integer matrix multiplication kernels to perform operations on quantized data directly. For int8 weights, it can leverage native tensor-core int8 GEMM. If a subsequent layer is also quantized, Quanto keeps the output in quantized form (stored as a special QTensor) to avoid unnecessary dequantization and requantization steps. Otherwise, the output of a quantized linear layer is computed in higher precision (int32) and then scaled back down. 

This quantized tensor propagation means that only a few key layers need to explicitly handle quantization boundaries. For example, a quantized Linear layer will take quantized input, multiply int8$\times$int8 $\rightarrow$ int32, apply the weight's scale and input scale to produce a correctly scaled output, and then pass along a quantized output tensor. Biases remain in full precision and are added after the integer matmul. Notably, Quanto supports float8 quantization on newer GPUs, using NVIDIA's FP8 formats for activations or weights when available (falling back to int8 if not).

\subsection{Distinctive Features and Tradeoffs}
Quanto extends PyTorch's native quantization with features like int2/int4 support (which core PyTorch lacks) and operation on any device (including GPU and Apple MPS). Unlike some LLM quantizers, Quanto does not fine-tune or adjust its quantization parameters using sample data to reduce the difference between the models quantized and original outputs. This makes it more general but potentially less accurate for very low-bit LLM quantization unless group-wise schemes are used. Quanto modifies standard PyTorch layers in-place, making it seamlessly compatible with existing models and workflows. In summary, Quanto provides a broad PTQ toolbox: any Linear/Conv layer can be quantized to 8, 4, or 2 bits with minimal code changes.

\section{HQQ (Half-Quadratic Quantization)}
HQQ \cite{hqq} is a data-free PTQ method that formulates weight quantization as an optimization problem. 

\subsection{Quantization Approach}
Unlike other strategies which require real calibration data to minimize discrepancies in layer outputs, HQQ directly minimizes reconstruction error in the weights, without relying on any input data. It assumes the pre-trained weights are the only input and tackles the quantization of each weight tensor by finding the optimal quantization parameters (scale and zero-point) that best reconstruct the original weights. HQQ's core idea is to make quantization more robust by reducing the impact of large magnitude outlier weights. It does this by modeling the quantization error with a heavy-tailed (hyper-Laplacian) loss function that penalizes large errors less aggressively, preventing a few extreme weights from dominating the optimization. Since solving this directly is complex, HQQ uses a two-step approach: one step adjusts the residual errors through a thresholding update that pushes small differences towards zero, and the other step updates the quantization parameters using a simple averaging rule. This alternating process leads to more stable and accurate low-bit quantizations. This procedure converges in only a few iterations (thanks to closed-form updates rather than slow gradient descent). 

HQQ supports a key parameter $axis$ which controls in which direction quantization groups are formed. Setting $axis=0$ means grouping along the output dimension so that each output neuron's weights are quantization separately. Conversely, setting $axis = 1$ groups along the input dimension, which usually matches hardware kernel layouts and runs faster, but can lead to a slight drop in accuracy since all outputs share the same quantization parameters. HQQ allows a flexible group\_size as well to balance granularity and storage overhead.

\subsection{Quantized Inference Mechanism}
The output of HQQ's optimization is a quantized model with integer weights and associated scales/zero-points for each group. Importantly, HQQ uses standard linear dequantization $\text{output} = \text{scale} \times (\text{int\_value} - \text{zero\_point})$, which means it is fully compatible with existing optimized inference kernels. Because HQQ quantization can be done for very low bit-widths, it leverages grouped or packed kernel implementations. HQQ itself does not enforce a custom runtime, producing quantized weights that can be loaded into a variety of backends. The implementation provides a wrapper HQQLinear module for PyTorch that replaces nn.Linear layers and under the hood unpacks the low-bit weights for multiplication. For maximum speed, HQQ recommends using axis=1 (per-column quant) in combination with fused CUDA/Triton kernels that can handle int4/int2 arithmetic. In fact, HQQ quantized models can be executed with existing high-performance kernels such as those from NVIDIA's CUTLASS or implementations like GPTQ's Triton kernel, since the dequantization is a linear scaling that can be fused into the matrix multiplication. If a less common configuration is chosen (e.g., per-row 4-bit), the library can fall back to a PyTorch implementation which performs dequantization in a slower loop on the fly. Aside from linear layers, due to its model-agnostic nature, HQQ can quantize other modules, but focus is usually placed on transformer linear layers since they dominate memory. HQQ stores its quantized weights as regular integers (int8 or smaller) for easy compatibility.

\subsection{Distinctive Features and Tradeoffs}
HQQ's major advantage is that it needs no calibration data at all; a user can quantize a model of any size in minutes just from its weights. This contrasts with GPTQ and AWQ which require sample data. It explicitly targets weight outliers (common in LLMs) whereas simpler quantization strategies might just clip or round them. HQQ supports a wide range of bit-widths (8, 4, 3, 2, and 1-bit) under a unified framework. Another difference is speed during the actual quantization phase: thanks to closed-form updates, HQQ can quantize extremely large models quickly (the authors report quantizing a 70B model to 2-bit in under 5 minutes). In summary, HQQ is a calibration-free, optimization-based quantizer that yields linearly quantized weights. It is often used as an on-the-fly quantizer (e.g., quantizing models at loading time) because of its speed and simplicity of deploying the resulting model.

\section{TorchAO} 
TorchAO \cite{torchao} is PyTorch's native library that includes advanced PTQ techniques, providing a high-level API to quantize models with minimal code changes.

\subsection{Quantization Approach}
TorchAO supports both traditional 8-bit quantization and newer 4-bit weight-only quantization for LLMs. The weight-only int4 mode applies quantization to nn.Linear (and similar) layers' weights while typically leaving activations in higher precision (FP16/FP32) to avoid accuracy loss. By default, it uses per-row quantization for linear weights: each output channel (row of the weight matrix) gets its own scale and zero-point (usually symmetric, so zero-point = 0) for 4-bit values. This row-wise granularity mitigates the variation in weight distributions across different neurons.

\subsection{Quantized Inference Mechanism}
Models quantized by TorchAO use custom module implementations and tensor subclasses to execute efficiently. When a linear layer is quantized, TorchAO replaces it with a special int4 kernel. Internally, weights are typically packed into 8-bit containers. For int4, each byte holds two 4-bit values, and for int8 no additional packing would occur. TorchAO's inference engine then uses low-level vectorized kernels to multiply quantized weights with the incoming activations. In an int4 weight-only scenario, the activations are still float16, so the kernel will read a block of int4 weights, unpack them into int8 or int16, multiply with the FP16 activation values, accumulate in FP16/FP32, and produce an FP16 output. (On NVIDIA GPUs, these kernels may use Tensor Cores by treating 4-bit weights as 8-bit by zero-extending). 

TorchAO emphasizes compatibility with torch.compile, meaning its quantized ops are written in a way that Torch's JIT or compiler can fuse them and optimize execution across modules. Under the hood, the quantized linear layer uses an operation sequence that follows three steps: (a) pack int4 weights into a contiguous int32 tensor, (b) on each forward pass, use a kernel to convert int4 to int16 on the fly and perform matrix multiplication with the FP16 inputs, (c) apply the per-row scale to the result. Because the per-row scales are constant, TorchAO can pre-multiply them into the weight values or post-multiply the output vector for each row. For int8 quantization, TorchAO can use more standard int8 * int8 GEMM routines. 

In addition, TorchAO's quantization is designed to integrate with sparsity: one advanced feature is the ability to impose 2:4 structured sparsity (meaning that out of every 4 consecutive weights, exactly 2 are non-zero and 2 are forced to zero) alongside int4 quantization, which further speeds up inference on supported hardware. Overall, during inference the layers operate mostly in low-precision space with minimal overhead from dequantization. Any necessary dequant (for operations that are not quantized, like a nonlinear activation or residual addition) is done as late as possible and quantization of outputs is done again before the next quantized layer. This mirrors the approach of PyTorch's static quantization, but TorchAO extends it to sub-8-bit regimes and GPU execution.

\subsection{Distinctive Features and Tradeoffs}
TorchAO is PyTorch's official framework for LLM low-bit quantization, so it places a strong emphasis on performance and integration. Unlike algorithm-specific quantizers (e.g., GPTQ, AWQ), TorchAO is not a standalone algorithm but a framework; it can incorporate techniques like HQQ or potentially others as options. This modular design means TorchAO can serve as a unified runtime for models quantized with different strategies. Compared to Hugging Face's Quanto, TorchAO is more focused on speed optimizations, whereas Quanto prioritizes broad support and ease of use. TorchAO's inclusion of FP8 dynamic quantization is forward-looking, anticipating new hardware capabilities. In contrast, most other tools focus only on int4/int8. In summary, TorchAO brings together many PTQ techniques under one roof, from straightforward int8 to cutting-edge int4 with grouping, backed by efficient kernels and optional enhancements like HQQ. It transforms a float model to a quantized model in one line, and ensures that the model can be run (and even fine-tuned with LoRA) just like a normal PyTorch module.

\section{Bitsandbytes} 
Bitsandbytes (bnb) \cite{bitsandbytes} provides lightweight quantization for LLMs with a focus on 4-bit and 8-bit weight compression. 

\subsection{Quantization Approach}
Bitsandbytes' 4-bit quantization method uses a novel NF4 (Normal Float 4) data type, introduced in the QLoRA paper \cite{qlora} to quantize weights nonlinearly. The process works as follows. First, each weight matrix is divided into fixed-size blocks (e.g., 64 values per block). For each block, bnb computes a scale factor equal to the block's maximum absolute value, $S = \max_{i}|w_i|$. The block's weights are then normalized by $S$. Next, instead of a linear quantization (which would map this range into 16 equal levels), Bitsandbytes uses a predefined set of 16 floating-point levels that are optimized for normally distributed data. These NF4 quantization levels are spaced such that there are more levels near 0 and fewer toward the extremes. In fact, they are chosen as quantiles of a standard normal distribution, then scaled. 

Each normalized weight in the block is assigned to the nearest of these 16 levels and the corresponding 4-bit code is stored. Thus, the weight block is compressed into 4-bit indices plus the scale $S$ needed for dequantization. This strategy effectively gives higher precision around 0, which is beneficial because neural network weights tend to cluster around zero. For 8-bit quantization, Bitsandbytes uses a simpler scheme where most weights are int8 but a few outlier columns stay in higher precision. Specifically, for int8 it identifies “outlier" feature dimensions (e.g., those with very large norms) and exempts them from quantization, storing them in 16-bit to preserve model accuracy. This avoids the extreme error that a single badly quantized column could introduce. Bitsandbytes does not typically require a calibration dataset, it relies on the weight distribution itself. It also offers an alternative FP4 quantization, but NF4 is recommended due to its better accuracy.

\subsection{Quantized Inference Mechanism}
At runtime, Bitsandbytes uses custom CUDA kernels to perform inference with quantized weights efficiently. The 4-bit weights are stored in a packed format where each byte contains two 4-bit values. The library provides implementations of a quantized matrix, vector multiply that unpack and dequantize on the fly. In essence, the computation is performed in higher precision after a partial dequantization, Bitsandbytes does not do pure integer arithmetic for 4-bit inference, it instead trades off memory for compute: storing weights at 4-bit, but doing arithmetic in 16-bit or 32-bit. The library allows configuring the computation dtype so that one can choose to accumulate in FP32 for maximal accuracy, or use FP16/BF16 for quicker compute. 

Bitsandbytes also implements a technique called double quantization: after quantizing weights to 4-bit with scales, it can further quantize the scale values themselves (since storing one FP16 scale per 64 weights still incurs memory). It quantizes those scales to 8 bits with a second-scale, reducing memory overhead of the scales. During inference, the double quantized scales are first dequantized to FP16 and then used to scale the weight blocks. This adds a minor overhead but saves memory. The net effect is that Bitsandbytes can load extremely large models in limited GPU memory by storing weights in 4-bit, and the runtime only modestly increases the compute (which is often bottlenecked by memory bandwidth rather than arithmetic). 

Additionally, Bitsandbytes provides optimized 8-bit kernels. For LLM.int8(), it splits each weight matrix into the int8 part and the high-precision part (outlier columns). The int8 part is multiplied efficiently meanwhile, the outlier columns (few in number) are multiplied in full-precision normally and added to the result. This way, the overhead of keeping some weights at full-precision is small but prevents large errors. Bitsandbytes integrates with Hugging Face Transformers via a bnb.MatrixMultiplicator or replacing linear layers with bnb.nn.Linear8Bit/Linear4Bit classes that override the forward pass. The rest of the model remains unchanged. One constraint is that Bitsandbytes currently quantizes only weights (activations remain in full-precision and are never quantized) to preserve model accuracy.

\subsection{Distinctive Features and Tradeoffs}
Bitsandbytes was one of the earliest tools to enable LLM inference at 8-bit and 4-bit without retraining, and it remains popular due to its simplicity. It does not minimize layer output error like GPTQ, nor does it use calibration data. It relies on statistical assumptions (e.g., weights follow normal distribution) and simple rounding. Therefore, purely on accuracy, it is not theoretically the most optimal design. However, Bitsandbytes's NF4 quantization is surprisingly effective. By using non-uniform levels, it retains much of the model accuracy. In terms of speed, Bitsandbytes runs on GPU but does not fully leverage int4 Tensor Cores, instead it uses half-precision operations. This is one reason libraries like EETQ (below) claim higher speed, EETQ uses true int8 cores, whereas bnb's 4-bit uses more instructions to unpack and multiply weights using a closed-form, fixed scheme. In summary, Bitsandbytes offers practical and memory-efficient PTQ by compressing weights with per-block scales, then performing inference by quick dequantization to higher precision on the fly. It is straightforward to use and does not need calibration, making it a common baseline for compressing large models.

\section{AutoAWQ (Activation-aware Weight Quantization)} 
AutoAWQ \cite{awq} is a library that implements AWQ, a PTQ strategy tailored for LLMs that uses a small calibration set to guide weight quantization. 

\subsection{Quantization Approach}
The primary insight of AWQ is that not all weights (or channels) are equally important for model accuracy. In particular, some weights (i.e., entire columns or rows of a weight matrix, corresponding to certain neurons) have a disproportionate effect on the output and are more sensitive to quantization error. AWQ identifies these salient channels by analyzing the activation distribution. It runs a set of representative inputs through the model and records statistics such as the variance or magnitude of each channel's outputs. For example, in a linear layer $Y = W X$, each output dimension $j$ has activation $y_j = w_j \cdot X$. If the distribution of $y_j$ has high variance or large magnitude relative to other channels, then the weights $w_j$ are “salient", a small change in them can cause a large change in output. AWQ's strategy is to protect these top channels during quantization. Instead of resorting to mixed-precision (which would leave those channels in higher bit-width, complicating hardware execution), AWQ uses an equivalent scaling transformation: it multiplies the salient weight channels by a factor to make them larger. To preserve the overall computation, it divides the corresponding activations by the same factor at inference time. 

In practice, this is done by folding the scale factor into the next layer's weights or the layer's normalization so that the model's outputs remain the same. The scale factors are determined offline from the collected activation stats. Essentially AWQ finds a scaling such that each protected channel's quantization error is below a certain threshold or that the channel's variance is brought in line with others. Typically, only a very small fraction of channels ($\sim$1\%) need significant scaling. After scaling, AWQ then quantizes all weights uniformly to low bit-widths. It uses standard per-channel linear quantization (with zero-point = 0) for all channels, including the scaled ones. Because the salient channels were amplified, when they are rounded to 4-bit integers the relative rounding error is much smaller. Thus, AWQ effectively achieves a result similar to mixed precision but within a single precision (e.g., all 4-bit), which is hardware-friendly and requires no special handling of different precisions. 

\subsection{Quantized Inference Mechanism}
The AutoAWQ library is an implementation that provides an efficient inference backend for AWQ-quantized models. AutoAWQ includes a custom CUDA kernel for 4-bit matrix multiplication, similar to other frameworks. It packs the int4 weights and uses a fused kernel to multiply the model's weights and activations. The kernel is optimized for both the context (prefill) stage and generation (decode) stage of LLM inference. It supports efficient batched GEMM for initial token processing and fast vector-matrix multiplies for single-token decoding. The result is that an AWQ-quantized model can run significantly faster and with less memory than the full-precision model. AutoAWQ specifically integrates with Hugging Face Transformers, so one can load a model that has been quantized with AWQ directly. It provides an AutoAWQForCausalLM class that will load the quantized weights. 

During the forward pass, the quantized linear layers call the AutoAWQ kernel. It reads 4-bit weight blocks, multiplies by input activations, and accumulates. Because AWQ quantization uses per-channel scales, the kernel must apply each channel's scale after the dot-product. A typical implementation is to perform the dot-product for one output channel completely, then multiply the result by that channel's scale. Since AWQ does not change activation precision, no special handling for activation quantization is needed. One extra step is the transformer layer norm, AWQ often relies on the model having a final layer normalization that can absorb the scaling differences. In some cases, AutoAWQ may tweak the model's normalization layers to account for the channel scaling (ensuring the outputs match the original model). But this is done once at model load time. In summary, inference with AWQ models is very similar to inference with plain 4-bit models (like GPTQ models), integer-packed weight multiplications plus a set of per-row scale factors applied to the output.

\subsection{Distinctive Features and Tradeoffs}
AWQ sits somewhere between simple data-free methods and complex calibration-based methods. It does use calibration data but in a lightweight manner, just to compute activation statistics, not to run heavy optimization. The method is fast to execute (scaling and rounding weights is O(N) per layer). Compared to GPTQ, AWQ is much faster in the quantization step (no iterative solving), and theoretically generalizes better to varied tasks because it does not overfit to a small calibration set through reconstruction. It primarily prevents catastrophic outlier errors. The AWQ paper noted that protecting as few as 1\% of channels was enough to greatly reduce the worst-case quantization error. This leads to strong results on benchmarks, especially for 4-bit. Another advantage is hardware-friendliness: all weights end up in int4, so inference can use one uniform kernel without branching for special cases. 

AutoAWQ, the library, further differentiates itself by providing integration with systems like Hugging Face and vLLM. In summary, AWQ's approach is activation-aware: by using knowledge of how weights are actually used (their activations), it achieves a good compromise of accuracy and efficiency. It is particularly noted for preserving performance on instruction-tuned and multi-modal models, where some other methods struggle.

\section{GPTQ (Generalized Post-Training Quantization)} 
GPTQ \cite{gptq} is an algorithm that treats quantization as a layer-wise error minimization problem.

\subsection{Quantization Approach}
For each layer (typically a large fully-connected weight matrix in an LLM), GPTQ optimizes the quantized weights to ensure the layer's output on representative inputs closely matches the original full-precision output. Unlike naive methods that round weights independently, GPTQ accounts for the correlations between weights through a process of sequential quantization with error compensation. The algorithm quantizes the weight matrix one group at a time. When a group is quantized, it inevitably introduces a small amount of error into the output. Before moving to the next column, the algorithm calculates this error and adjusts the remaining unquantized weights to compensate for it. This update relies on an approximate Hessian to determine exactly how the remaining weights should be shifted to absorb the error, ensuring the total degradation of the layer's output is minimized.

Conceptually, it means if quantizing column $j$ caused a small output difference, you nudge the other columns' weights in the opposite direction to cancel out that difference. The algorithm thus alternates between quantizing one column (find the best quantized values for that column given current residual), then recompute the residual on $Y$ and apply a correction to all subsequent float columns to absorb the error. To perform these steps efficiently, GPTQ precomputes some matrices. In the literature, they derive a formula using Cholesky decompositions to solve for the optimal $q_j$ in closed form. The result is that GPTQ produces a quantized weight matrix that is extremely close to the original in terms of output on the calibration set. It typically uses asymmetric quantization for weights (meaning a zero-point may be non-zero) to better fit the distribution. For LLMs, GPTQ is commonly run with 4 bits per weight and a small calibration set (e.g., 128 random samples or a few hundred tokens from an evaluation dataset). One can also configure a group size instead of quantizing entire columns at once. This introduces more scales/zero-points (one per subgroup) and further reduces error, at the cost of extra storage for those scales. In many implementations, group\_size=128 or group\_size=64 is used as a sweet spot for 4-bit quantization of large models, it cuts down error significantly compared to per-column quantization.

\subsection{Quantized Inference Mechanism}
A model quantized with GPTQ ends up with weight matrices in a custom format: for each weight matrix, there are typically: a set of small int matrices (each storing the quantized values for a group of weights) and accompanying scale/zero-point vectors for each group. For example, a linear layer of shape [out\_dim, in\_dim] quantized to 4-bit with group\_size 128 will be stored as $\lceil\frac{\text{in\_dim}}{128}\rceil$ groups per output. Each group has a 4-bit weight matrix of shape [out\_dim, 128] and a scale and zero-point per out\_dim (if grouping along input axis) or per 128 block segment. At inference, these quantized weights are used much like in Bitsandbytes or AWQ. For each output channel, the kernel multiplies 4-bit weights by full-precision inputs, accumulates in FP32, then multiplies by the weight scale and adds bias. If a zero-point (offset) is used, the kernel subtracts the zero-point during the multiplication (effectively shifting the integer values up or down before scaling). Because GPTQ often uses per-row scales, the scaling can be applied as a simple vector multiply to the output of the matmul. In summary, inference looks exactly like a standard affine quantized layer, done with efficient low-bit arithmetic. Many GPTQ-quantized models can run at nearly the same speed as 8-bit models, since the bottleneck is memory bandwidth and 4-bit cuts memory movement in half.

\subsection{Distinctive Features and Tradeoffs}
By explicitly minimizing output errors and using calibration data, GPTQ outperforms naive methods in preserving model quality. However, this comes at the cost of complexity: quantization with GPTQ is relatively slow (it can take hours for very large models) and requires storing the calibration data's activations (it precomputes $X X^T$ or similar, which can be heavy for big layers). HQQ, by contrast, is much faster but might not always reach the same level of accuracy on extremely low bits because it ignores output correlations. AWQ focuses only on a subset of weights and does not account for interactions between weights, so GPTQ often gives lower final error than AWQ as well. Another point is that GPTQ quantizes one layer at a time, assuming others remain in float during that layer's calibration. This layer-wise independence simplifies the process but is not globally optimal. In practice, 4-bit GPTQ has become a common standard for compressing models like LLaMA 7B,70B with minimal performance loss. Finally, GPTQ's outputs are often used as input to further fine-tuning, showcasing that it preserves model functionality well. 

In summary, GPTQ is an output-error-minimizing quantizer that produces quantized weights (usually groupwise int4) ready for fast inference. Its custom kernels and data structures are a bit more complex than uniform quantizers, but numerous libraries now make it seamless to load and use GPTQ-quantized models. This work demonstrates that PTQ can sometimes push to 4 bits without significant accuracy loss, using clever error compensation instead of more brute-force methods.

\section{EETQ (Easy \& Efficient Transformer Quantization)}
EETQ \cite{eetq} is a quantization strategy that focuses on 8-bit weight-only quantization for transformer models.

\subsection{Quantization Approach}
EETQ offers a relatively straightforward approach to quantizing all linear and embedding weights to int8 in a per-channel manner, similar to symmetric per-channel min-max quantization used in other quantization frameworks. The zero-point is usually set to 0. Because EETQ targets 8-bit (which has 256 levels) and most transformer weight distributions are reasonably well-behaved, this linear quantization already achieves high accuracy, typically incurring very small relative errors. Importantly, no calibration data is required for basic EETQ quantization, instead relying on its own weights. EETQ quantizes weights on the fly when loading a model, allowing the user to provide a full-precision model and an EetqConfig("int8") to the loader, it converts all appropriate matrices to int8 with scales. EETQ does not quantize activations, it assumes activations will remain unchanged at runtime.

\subsection{Quantized Inference Mechanism}
EETQ is paired with the EET (Easy \& Efficient Transformer) inference engine, which is highly optimized for NVIDIA GPUs. After quantization, each linear layer has int8 weights and a per-row scale vector. To compute $y = W_q x$ (where $W_q$ are int8 and represent approximately $W$), EETQ uses custom fused GPU kernels that leverage NVIDIA's Tensor Cores in int8 mode. On modern GPUs, Tensor Cores can perform 8-bit integer matrix multiply-accumulate very quickly, but they typically multiply int8 * int8 (giving int32 accumulations). Dynamic quantization of the input can be as simple as scaling the input vector by a constant factor (to use the full int8 range) each time. The phrase “per-channel weight-only quantization" means the output of each channel will be scaled correctly by the weight's scale. The result is that EETQ quantized models run extremely fast, since they effectively use int8 arithmetic throughout.

\subsection{Distinctive Features and Tradeoffs}
EETQ is a more conservative quantization scheme compared to others in this presented list of quantization strategies (since it stays at 8-bit). The benefit is that accuracy loss is minimal and well within what can often be recovered by subsequent fine-tuning if needed. EETQ's primary focus is on speed and efficiency. By quantizing to int8 and using optimized kernels, it significantly boosts throughput. In many deployment scenarios, int8 quantization (with good kernels) is ideal for production because it retains full accuracy and gains some speed. 4-bit methods might save more memory but can suffer slight accuracy drops or lack stable hardware support. Because it is weight-only, one can still use techniques like LoRA fine-tuning on top of an EETQ model (LoRA layers remain in full-precision and are added to the int8 weights during inference). The Hugging Face documentation explicitly mentions LoRA fine-tuning is supported on EETQ quantized models. 

In comparison to PyTorch's native int8 quantization which is mainly CPU-focused and might not fully exploit GPU tensor cores, EETQ is specialized for GPU. EETQ was built to be simple, exposing a single knob (“int8") to the user, and letting the library handle the rest, whereas something like TorchAO or Quanto provide more configuration options. In summary, EETQ quantization uniformly quantizes all weights to int8, uses efficient per-channel scaling, and relies on a highly optimized transformer engine to execute the int8 model. This yields a noticeable throughput improvement and memory reduction, with negligible accuracy loss, making it attractive for enterprise deployments that need to serve LLMs at scale.

\chapter{Evaluation}

This section presents a comprehensive evaluation of energy efficiency, runtime performance, and memory usage across a range of quantized LLM configurations.

\section{Methodology}
We benchmark five popular, open-source models representative of state-of-the-art transformer architectures: AllenAI OLMo 1B \cite{allenai}, Llama 2 7B \cite{llama2}, Phi 3.5 Mini Instruct (3.8B) \cite{abdin2024phi}, Qwen 2.5 7B \cite{qwen2.5}, and Qwen 2.5 14B. Most transformer architectures at their core consist mainly of matrix multiplication operations, which are impacted by quantization techniques the most. We evaluate these models across 4 tasks with varying output lengths. GLUE MNLI \cite{glue} and MMLU \cite{mmlu}, which involve natural language inference and multiple-choice question answering respectively, generate very short single-token outputs, while HumanEval \cite{humaneval} and GSM8K \cite{gsm8k} require long-form text generation, including multi-line Python functions and step-by-step math solutions. For GSM8K, we evaluate a 10\% sample of the dataset due to the high computational cost and the low variance observed in accuracy outcomes.

We evaluate ten quantization strategies derived from seven distinct libraries, including AutoAWQ int4, Bitsandbytes nf4, Bitsandbytes int8, EETQ int8, GPTQ int4, HQQ int4, HQQ int8, Quanto int4, Quanto int8, TorchAO int4, and compare a full-precision FP32 baseline. These methods vary substantially in their backend implementations, ranging from fused CUDA A100 and Triton \cite{triton} kernels to PyTorch matrix multiplication paths, which affect both overhead and scalability. When initializing these quantization strategies we use the default settings, which is symmetric quantization for all strategies, noting that their granularities differ: GPTQ, HQQ, and TorchAO default to per-group quantization; Bitsandbytes uses per-group for int4 and per-channel quantization for int8; and Quanto and EETQ use per-channel quantization. None of the evaluated strategies use pure per-tensor quantization by default.

Experiments were conducted over four NVIDIA GPUs: the A100, H100, and RTX A6000, representing higher-end datacenter accelerators, and the L4, a low-power inference-optimized GPU. Inference was performed using the PyTorch framework and the Hugging Face Transformers library. We deliberately avoided specialized inference engines such as vLLM \cite{vllm} or NVIDIA TensorRT-LLM \cite{tensorrtllm}, as their integrated optimizations would conflate the performance gains of the orchestration framework with those of the quantization methods. This choice ensures the isolation of quantization-induced gains, providing a controlled environment to measure the direct impact of each quantization strategy.

Metrics collected include total energy consumption, runtime, memory usage, and accuracy degradation relative to the vanilla FP32 baseline, which refers to the unquantized full-precision (32-bit float) model without any compression or optimization. This choice aligns with prior studies such as Shen et al.~\cite{fp32survey}, who evaluate 8-bit formats for post-training quantization across a diverse set of network architectures, using FP32 as their primary comparison point. Moreover, FP32 remains the golden standard for accuracy, while FP16 introduces numerical approximations that can already diverge from the original model fidelity. Since our goal is to isolate the effects of quantization, using FP32 avoids conflating quantization-induced error with half-precision optimization effects. This provides a clearer lens through which to analyze the trade-offs between accuracy, memory footprint, runtime, and energy consumption introduced by quantization strategies alone. For all results presented as a ratio (Energy Efficiency, Speedup, and Memory Reduction), we normalize the performance of the vanilla FP32 model to 1.0. Ratios greater than 1.0 indicate an improvement over the baseline.

To ensure that energy measurements reflect only the workload under evaluation, each benchmark was run on an otherwise idle GPU. Before each experiment, we verified that no processes were active on the device using the NVIDIA System Management Interface (nvidia-smi). GPU energy consumption was then measured using the ZeusMonitor interface \cite{zeus}, which records energy in joules over user-defined execution windows. For each run, ZeusMonitor tracked GPU power draw throughout the inference period and returned total energy and elapsed time. This setup provides a consistent and reproducible measurement of GPU energy usage across all quantization strategies.

In the following subsections, we analyze how quantization strategies impact energy efficiency across different output lengths, and discuss how kernel design and runtime characteristics shape real-world performance.

\section{Energy Efficiency \& Speedup}
LLMs are autoregressive models that generate text one token at a time, requiring substantial computation in their self-attention layers \cite{attention} for each new token.
To reduce the cost of computing self-attention, modern systems use key-value (KV) caching \cite{kvcache}: storing the key and value vectors from prior self-attention computations, so that they can be reused for each new token.

During the input phase, each attention layer computes attention scores via matrix multiplication
\[
   \mathbf{Q} \mathbf{K}^\top
\]
where
\begin{itemize}
    \item $\mathbf{Q} \in \mathbb{R}^{L \times d_\mathrm{model}}$ is the queries matrix,
    \item $\mathbf{K} \in \mathbb{R}^{L \times d_\mathrm{model}}$ is the keys matrix,
    \item $L$ is the input sequence length (number of tokens),
    \item $d_\mathrm{model}$ is the model's hidden size (e.g., $4096$ in Llama 2 7B).
\end{itemize}

This multiplication involves $(L \times d_\mathrm{model}) \times (d_\mathrm{model} \times L)$, resulting in a $(L \times L)$ attention score matrix. The computational cost thus scales quadratically with input length $L$.

In the decode (autoregressive generation) phase, the model attends to previously generated tokens one at a time with
\[
    \mathbf{q} \mathbf{K}_\mathrm{past}^\top
\]
where
\begin{itemize}
    \item $\mathbf{q} \in \mathbb{R}^{1 \times d_\mathrm{model}}$ is the query vector for the current token,
    \item $\mathbf{K}_\mathrm{past} \in \mathbb{R}^{L_\mathrm{past} \times d_\mathrm{model}}$ are keys for the past context,
    \item $L_\mathrm{past}$ is the number of tokens in the previously generated context.
\end{itemize}
Here, the multiplication is $(1 \times d_\mathrm{model}) \times (d_\mathrm{model} \times L_\mathrm{past})$, resulting in a $(1 \times L_\mathrm{past})$ vector. This operation scales linearly with the context length $L_\mathrm{past}$.

\begin{figure}[htbp]
\vspace{-1em}
\centering
\begin{subfigure}{0.8\linewidth}
  \includegraphics[width=\linewidth]{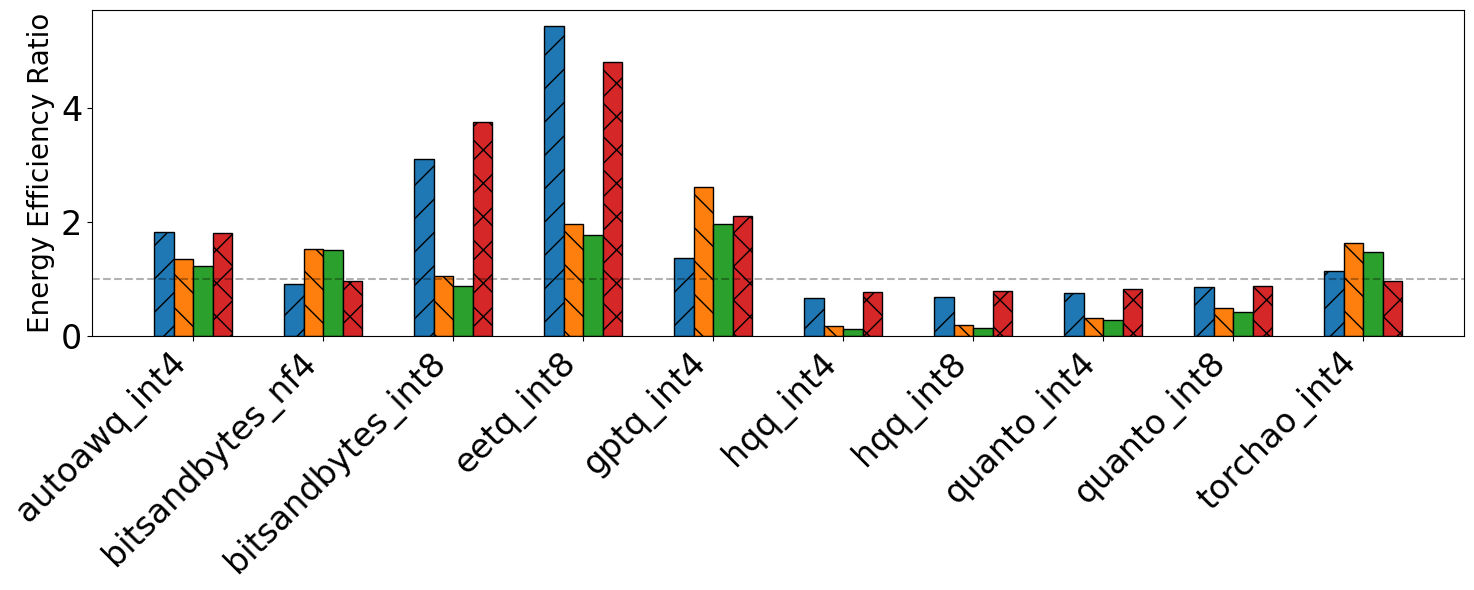}
  \caption{Energy Efficiency Ratio for Llama 2 7B}
\end{subfigure}
\begin{subfigure}{0.8\linewidth}
  \includegraphics[width=\linewidth]{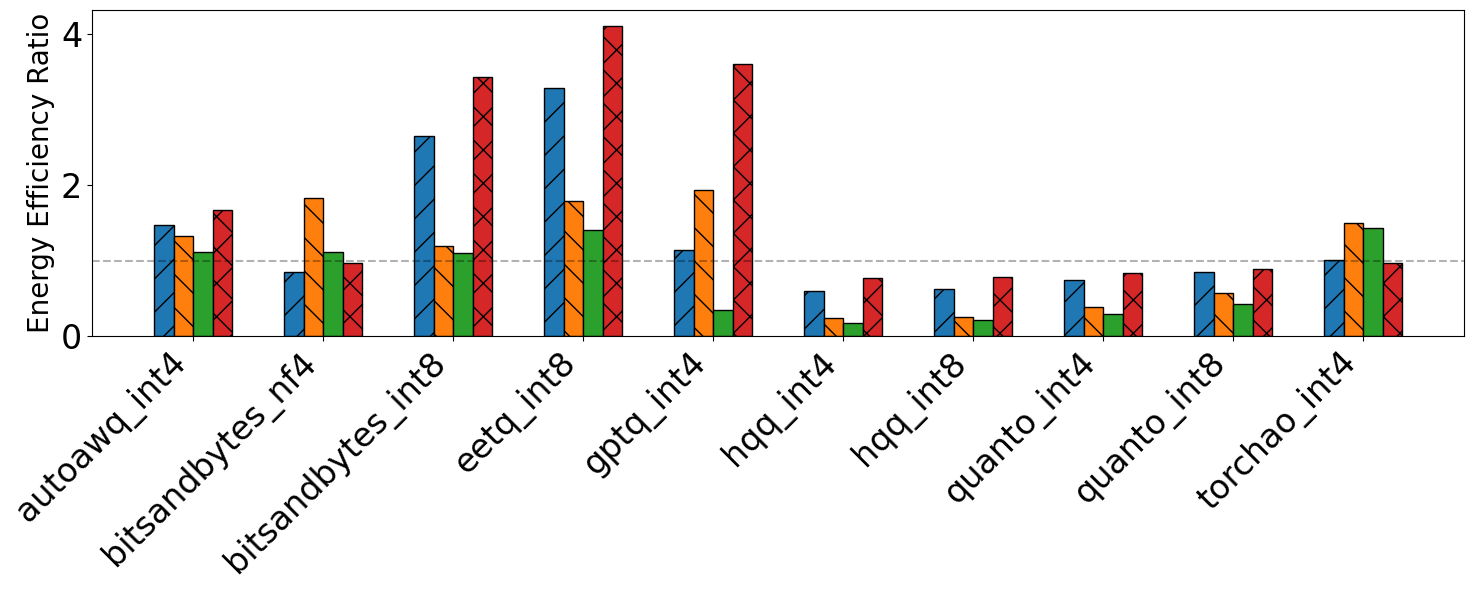}
  \caption{Energy Efficiency Ratio for Phi 3.5 Mini Instruct}
\end{subfigure}
\begin{subfigure}{0.8\linewidth}
  \includegraphics[width=\linewidth]{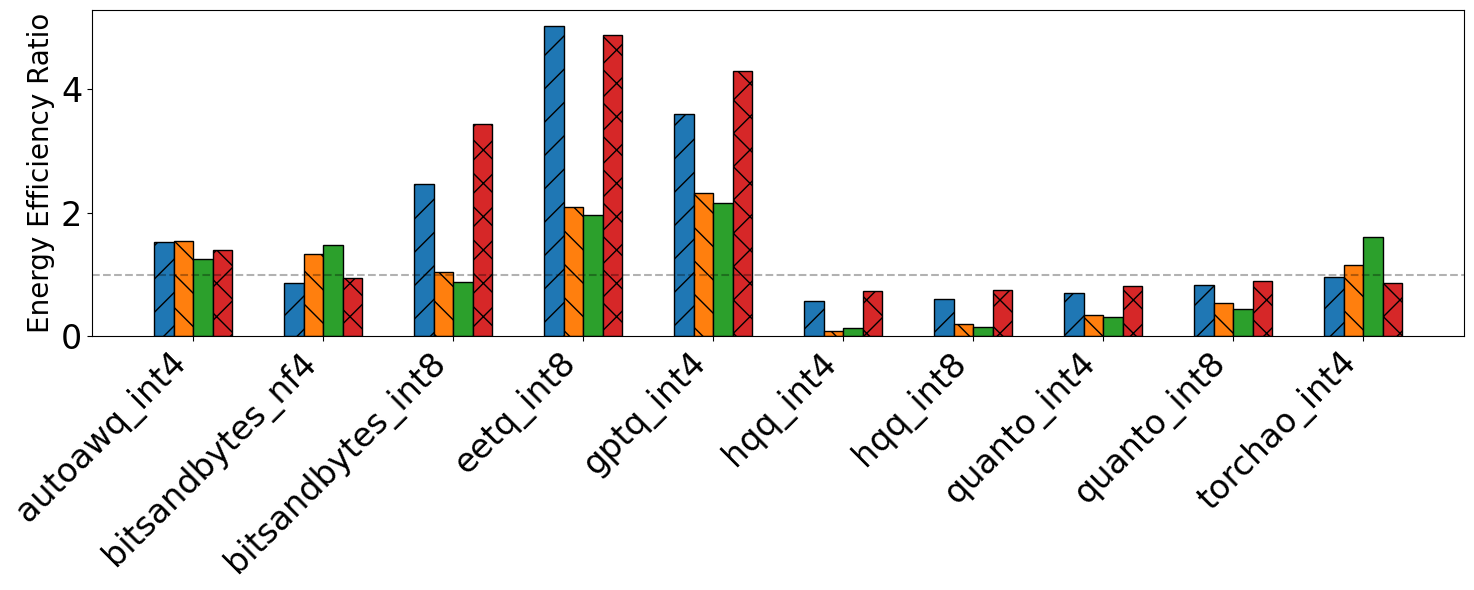}
  \caption{Energy Efficiency Ratio for Qwen 2.5 7B}
\end{subfigure}
\includegraphics[width=0.8\linewidth, trim=0 20 0 20, clip]{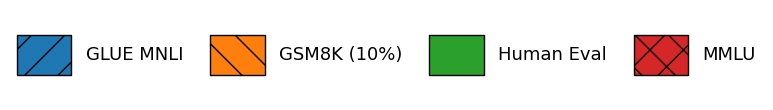}

\caption{Energy Efficiency Ratio Across Models and Datasets on A6000}
\label{fig:benchmarksA6000}
\vspace{-0.2cm}
\end{figure}

Figure~\ref{fig:benchmarksA6000} presents the energy efficiency ratios (i.e., baseline FP32 energy use divided by quantized energy use) for various quantization strategies on an Nvidia A6000 GPU, revealing distinct patterns and clear winners. In particular, the int8 methods - EETQ int8 and Bitsandbytes int8 - consistently achieve the highest efficiency ratios across all three tested models (Llama 2 7B, Phi 3.5 Mini, and Qwen 2.5 7B). GPTQ int4 also performs well, notably on GSM8K and HumanEval which involve longer outputs. These methods yield energy savings significantly above the 1.0 baseline, demonstrating the substantial energy reductions they offer relative to the full-precision baseline. It is interesting to note that Quanto int4/int8, EETQ int8, and Bitsandbytes int8 are unique among the tested strategies in their use of per-channel quantization instead of per-group quantization, with EETQ int8 and Bitsandbytes int8 performing the best. This might suggest that per-channel quantization, despite its potential for increased metadata overhead, offers advantages in capturing nuanced weight distributions and enabling efficient low-bit kernel implementations that, in turn, lead to strong energy efficiency. However, these gains should be interpreted as arising from the combination of per-channel quantization and optimized kernel implementations, rather than quantization granularity in isolation.

In Figure~\ref{fig:benchmarksH100} we extended these experiments on a top of the line Nvidia H100 GPU, as well as a smaller 1B model and a larger 14B model to show generalization. Unfortunately EETQ is not yet supported on H100 architecture ~\cite{not_supported}, but the rest of the quantization models show the same findings as the A6000 energy efficiency experiments, demonstrating that these trends can generalize well.

Other quantization strategies, including AutoAWQ int4, Bitsandbytes nf4, and TorchAO int4, also improve energy efficiency, but inconsistently across benchmarks and to a lesser extent due to the limited native acceleration for 4-bit arithmetic on current hardware. These int4 quantizers introduce additional overhead or underutilize tensor core throughput, which can reduce the expected benefits, aligning with reports that quantization without specialized hardware support can even slow down inference in some cases~\cite{survey1}. 

Figure~\ref{fig:energy_vs_output_length} (legend and caption on page~\pageref{fig:quant_strat_leg}) shows energy efficiency ratios for autogenerated randomized data when input lengths are fixed and output lengths vary. The trends observed show that quantization implementations accelerate large matrix multiplications (matmuls) more effectively than smaller ones. In particular, during the prefill phase, where large attention matrices $(L \times L)$ are processed, low-bit quantized matmuls result in substantial energy improvements over vanilla FP32. However, in the decode phase, where matrix multiplications are smaller (for example, $1 \times L$), the relative energy advantage diminishes. This is partly because the overall runtime for these smaller operations is lower, so the fixed overheads associated with quantization such as kernel launch costs and memory management account for a larger proportion of the total runtime, thereby limiting the potential energy savings. Consistent with these results, our benchmarks for MMLU and GLUE MNLI, which generate only single-token outputs, show substantial reductions in energy consumption, whereas benchmarks HumanEval and GSM8K that involve long-form text or code generation exhibit minimal energy savings due to their decode-heavy workloads. 

It is interesting to note that when input sizes are large (such as 1000 or 5000 tokens), prefill dominates the computation, and quantized models achieve considerable energy savings, often exceeding a 4 to 8 times improvement over vanilla FP32. On the other hand, when input sizes are small (for instance, 10 tokens), decode costs dominate, and the energy efficiency advantage of quantization is reduced. These efficiency gains are observed only when the prefill matmuls are sufficiently large. For example, on Llama 2 7B running on an A6000, gains begin to appear around 100 tokens, whereas on a higher-memory GPU like the A100, significant improvements emerge only around 5000 tokens.

\begin{figure}[htbp]
\vspace{-1em}
\centering
\begin{subfigure}{0.8\linewidth}
  \includegraphics[width=\linewidth]{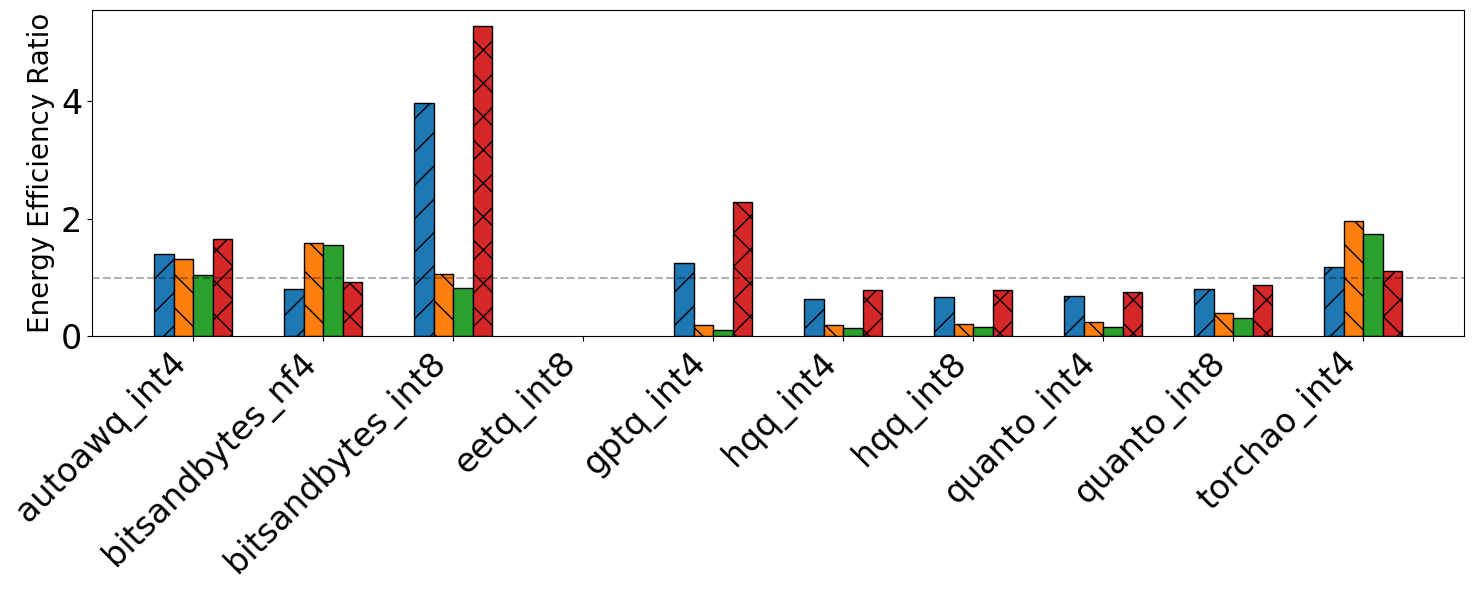}
  \caption{Energy Efficiency Ratio for Llama 2 7B}
\end{subfigure}
\begin{subfigure}{0.8\linewidth}
  \includegraphics[width=\linewidth]{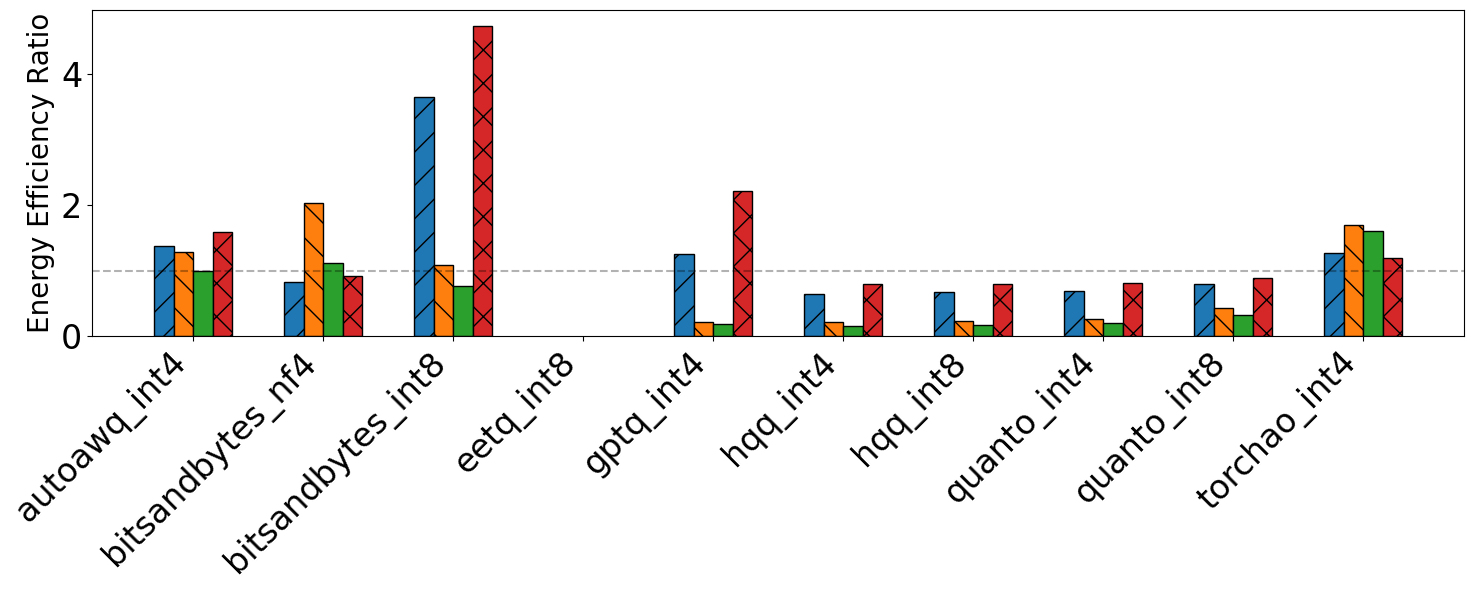}
  \caption{Energy Efficiency Ratio for Phi 3.5 Mini Instruct}
\end{subfigure}
\begin{subfigure}{0.8\linewidth}
  \includegraphics[width=\linewidth]{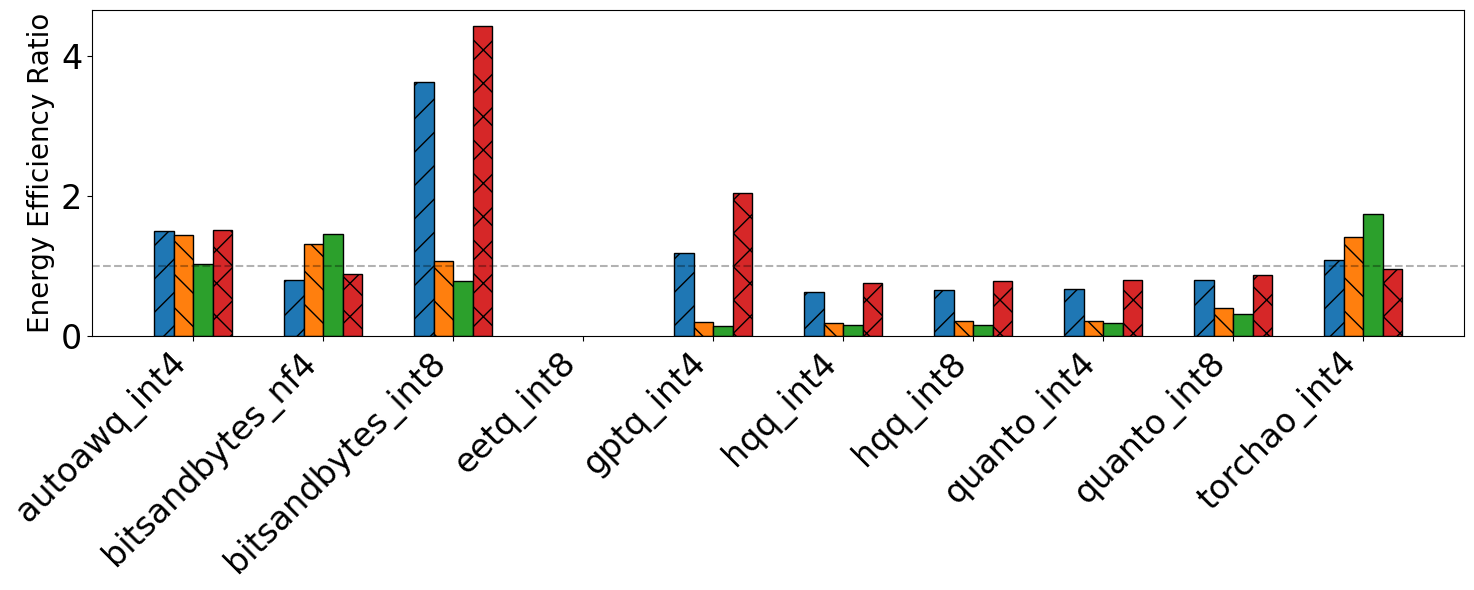}
  \caption{Energy Efficiency Ratio for Qwen 2.5 7B}
\end{subfigure}

\vspace{-0.2cm}
\end{figure}

\begin{figure}[htbp]\ContinuedFloat
\centering
\begin{subfigure}{0.8\linewidth}
\includegraphics[width=\linewidth]{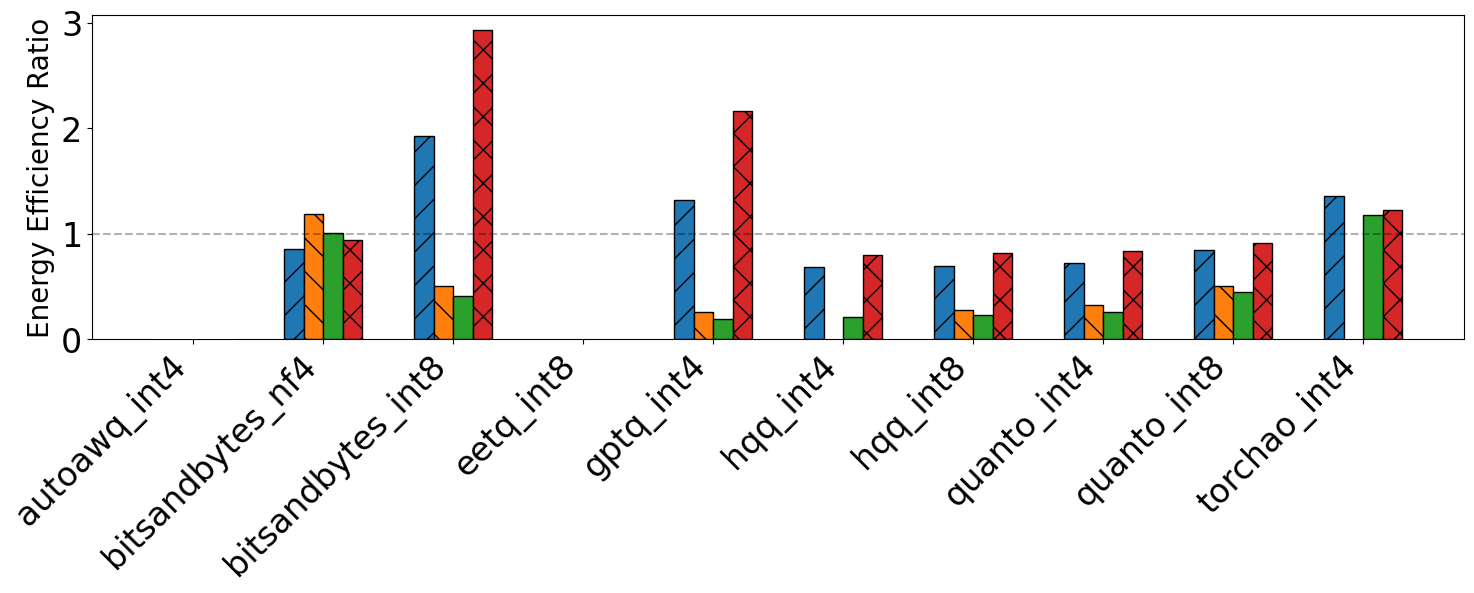}

  \caption{Energy Efficiency Ratio for OLMo 1B}
\end{subfigure}
\begin{subfigure}{0.8\linewidth}
  \includegraphics[width=\linewidth]{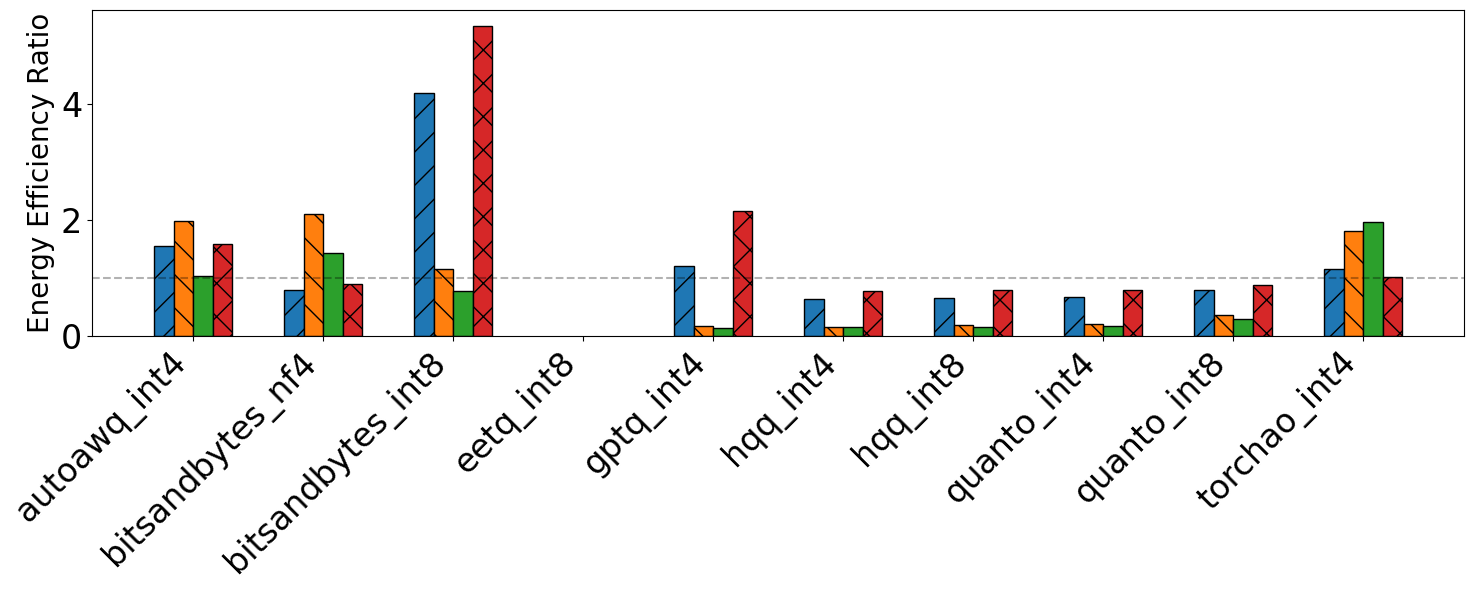}
  \caption{Energy Efficiency Ratio for Qwen 2.5 14B}
\end{subfigure}

\includegraphics[width=0.8\linewidth, trim=0 0 0 0, clip]{Images/Benchmark_Legend_Flat.png}
\caption{Energy Efficiency Ratio Across Models and Datasets on H100}
\label{fig:benchmarksH100}
\end{figure}

\begin{figure}[htbp]
\centering

\begin{subfigure}{0.6\linewidth}
  \centering
  \includegraphics[width=\linewidth]{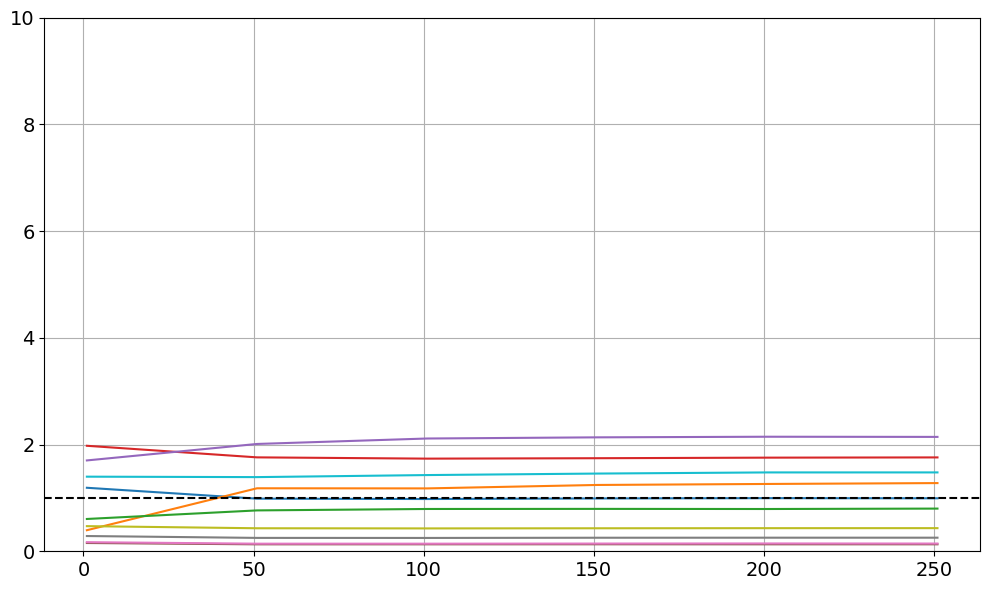}
  \caption{Input = 10, Llama 2 7B on A6000}
\end{subfigure}\par\medskip

\begin{subfigure}{0.6\linewidth}
  \centering
  \includegraphics[width=\linewidth]{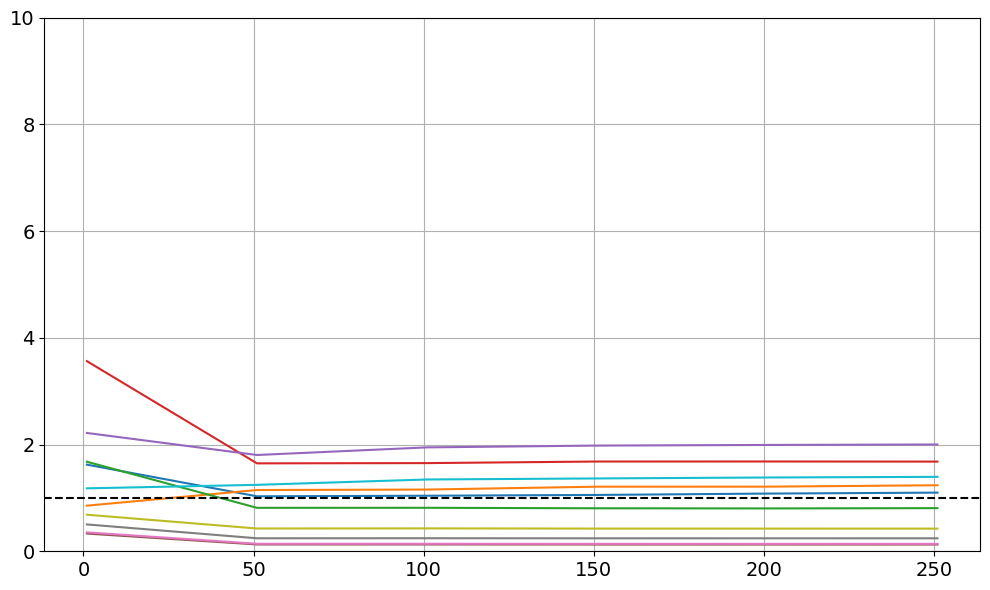}
  \caption{Input = 100, Llama 2 7B on A6000}
\end{subfigure}\par\medskip

\begin{subfigure}{0.6\linewidth}
  \centering
  \includegraphics[width=\linewidth]{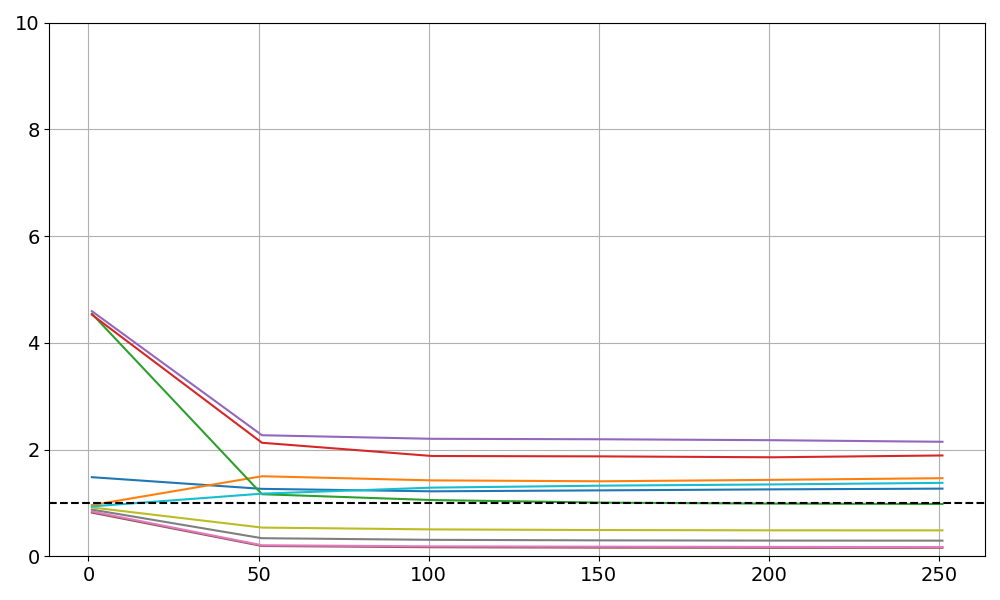}
  \caption{Input = 1000, Llama 2 7B on A6000}
\end{subfigure}

\end{figure}

\begin{figure}[htbp]\ContinuedFloat
\centering

\begin{subfigure}{0.6\linewidth}
  \centering
  \includegraphics[width=\linewidth]{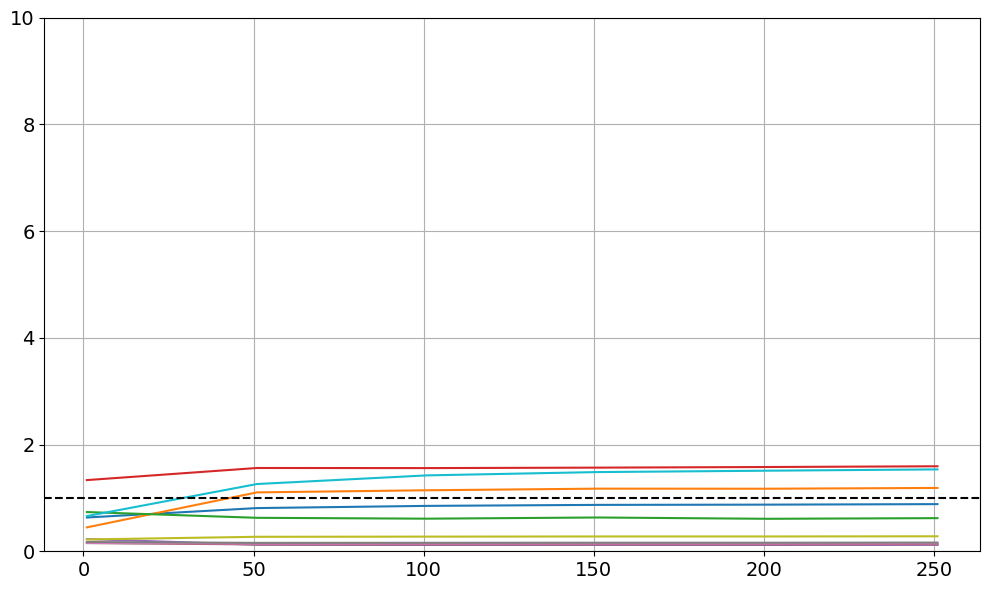}
  \caption{Input = 100, Llama 2 7B on A100}
\end{subfigure}\par\medskip

\begin{subfigure}{0.6\linewidth}
  \centering
  \includegraphics[width=\linewidth]{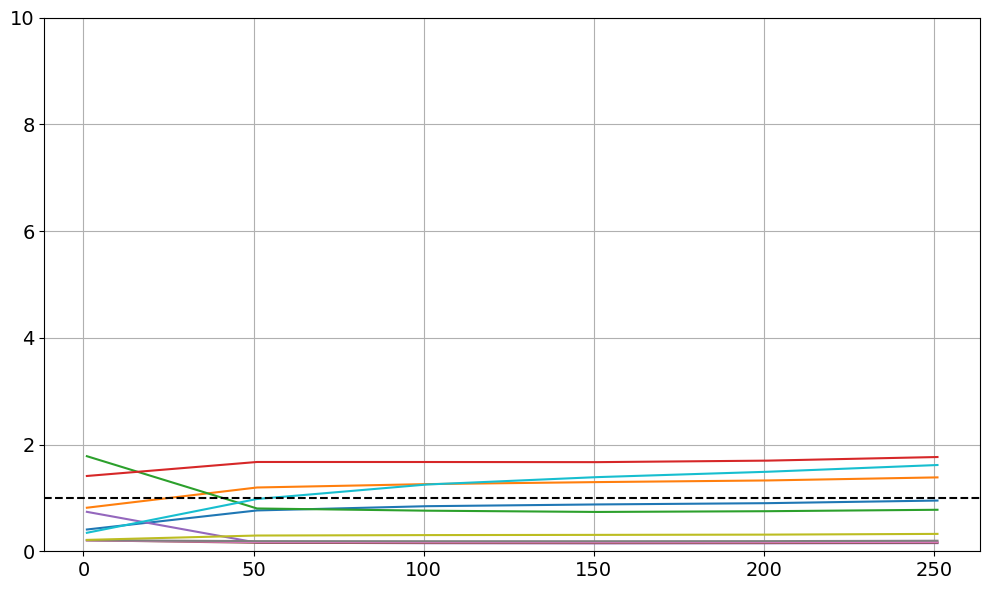}
  \caption{Input = 1000, Llama 2 7B on A100}
\end{subfigure}\par\medskip

\begin{subfigure}{0.6\linewidth}
  \centering
  \includegraphics[width=\linewidth]{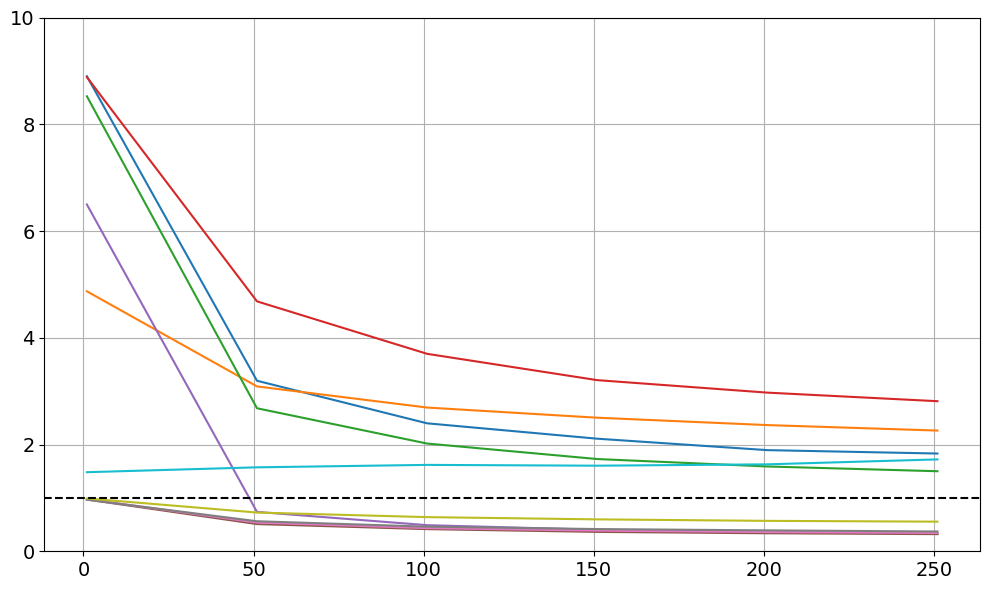}
  \caption{Input = 5000, Llama 2 7B on A100}
\end{subfigure}

\end{figure}

\begin{figure}[htbp]\ContinuedFloat
\centering

\begin{subfigure}{0.6\linewidth}
  \centering
  \includegraphics[width=\linewidth]{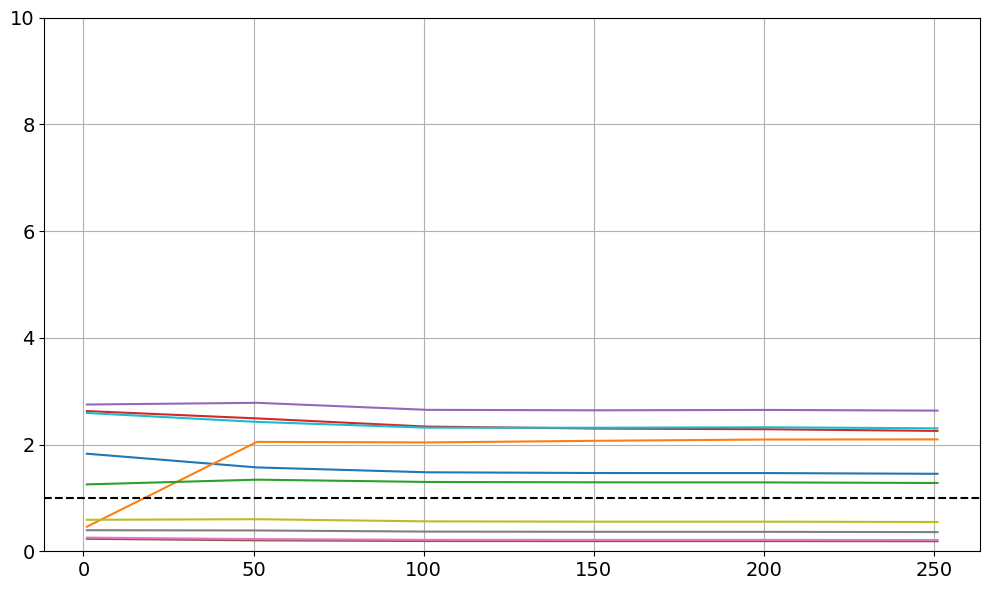}
  \caption{Input = 10, Phi 3.5 Mini Instruct on L4}
\end{subfigure}\par\medskip

\begin{subfigure}{0.6\linewidth}
  \centering
  \includegraphics[width=\linewidth]{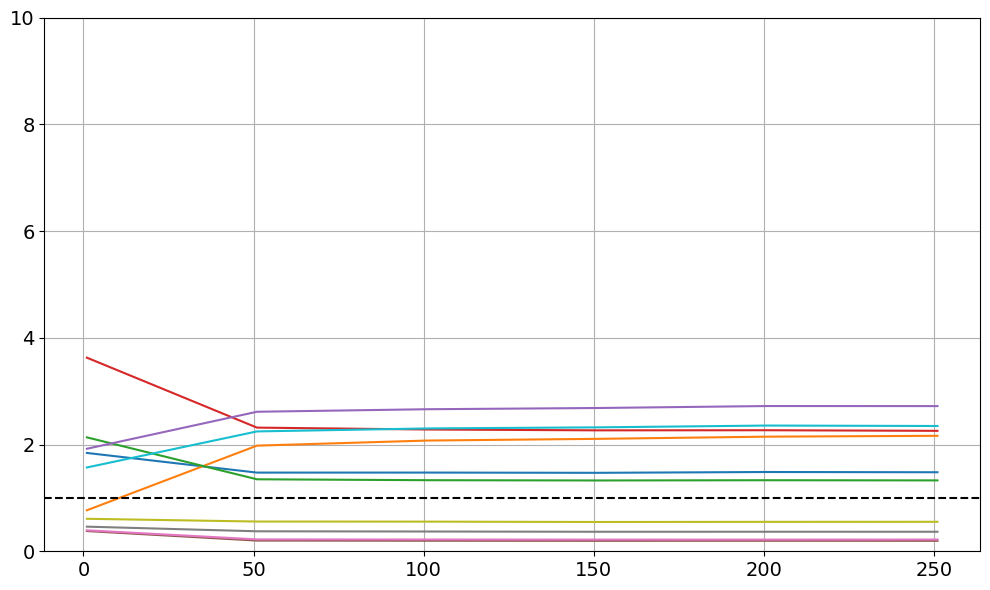}
  \caption{Input = 100, Phi 3.5 Mini Instruct on L4}
\end{subfigure}\par\medskip

\begin{subfigure}{0.6\linewidth}
  \centering
  \includegraphics[width=\linewidth]{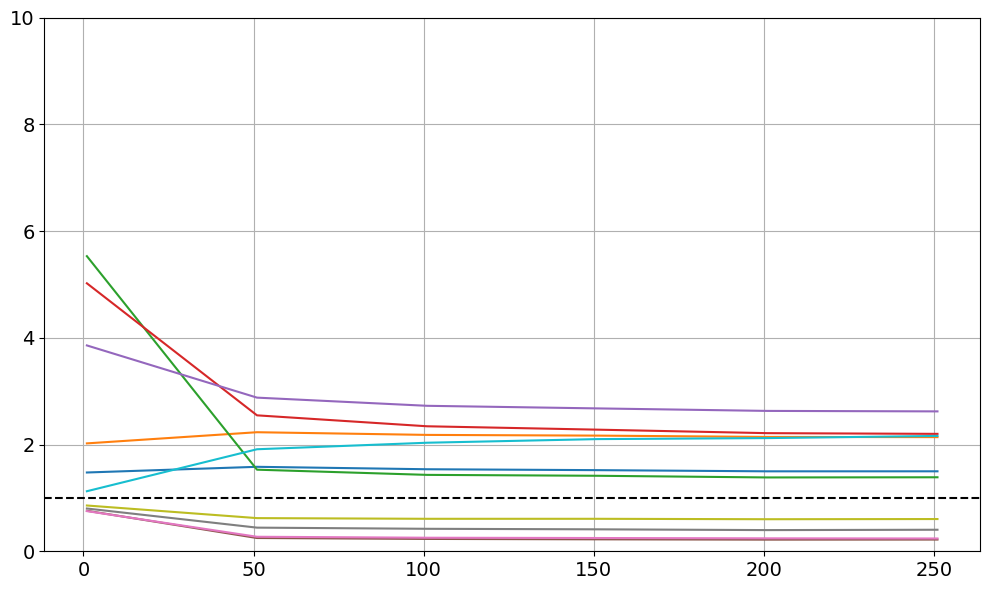}
  \caption{Input = 1000, Phi 3.5 Mini Instruct on L4}
\end{subfigure}
\end{figure}

\begin{figure}[htbp]\ContinuedFloat
\centering

\begin{subfigure}{\linewidth}
  \centering
  \includegraphics[width=\linewidth, trim=0 100 0 0, clip]{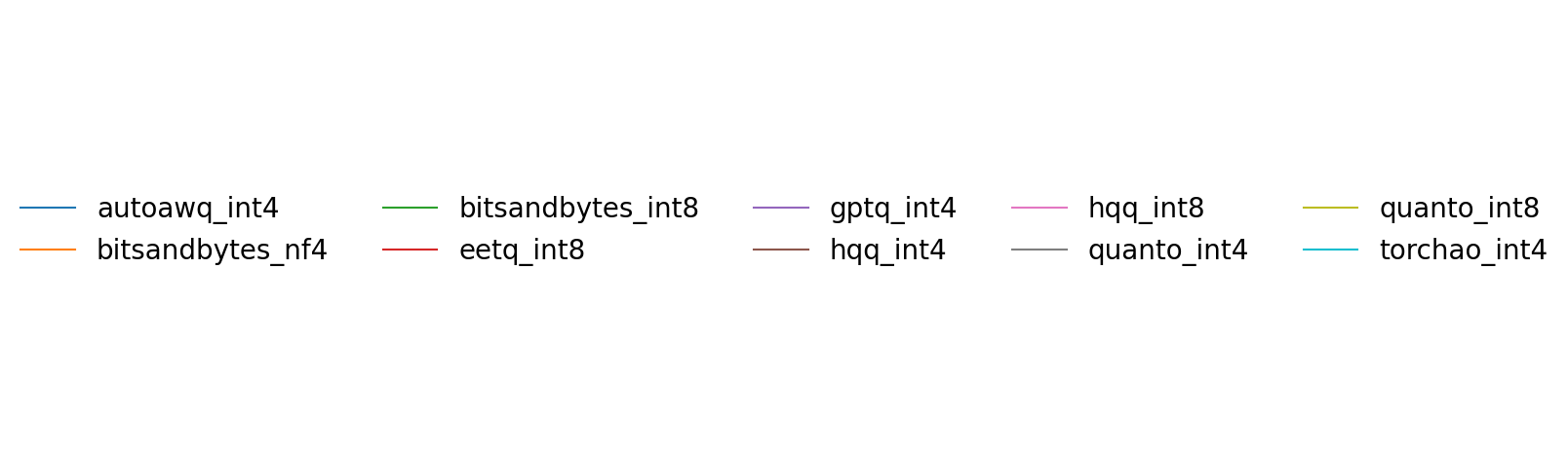}
  \label{fig:quant_strat_leg}

\end{subfigure}

\caption{Energy Efficiency Ratio vs. Output Length Across Different Input Sizes, Models, and Hardware}
\label{fig:energy_vs_output_length}
\end{figure}

\begin{figure}[htbp]
  \centering
  \begin{subfigure}[b]{0.75\linewidth}
    \centering
    \includegraphics[width=\linewidth]{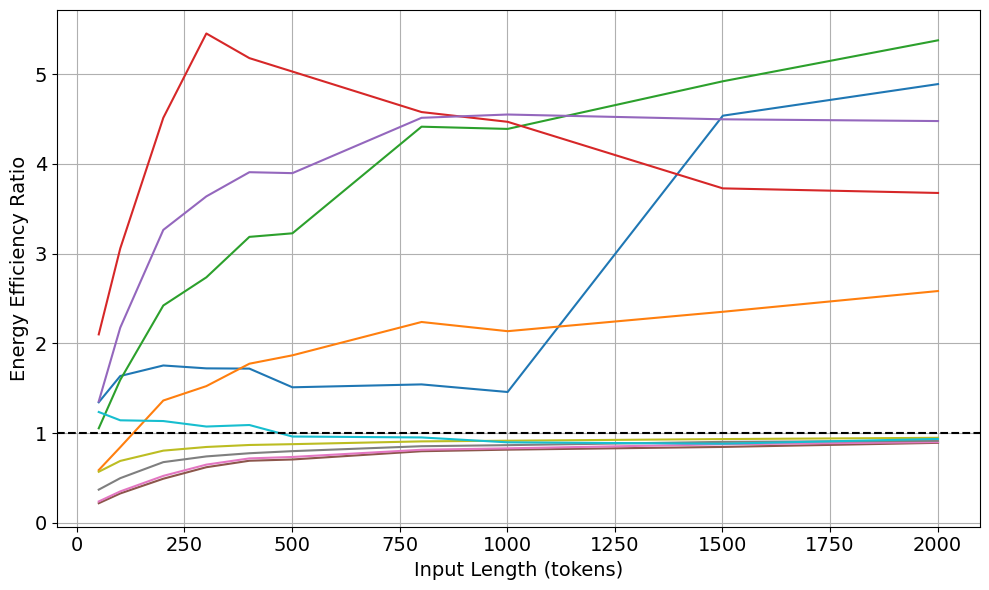}
    \caption{A6000}
  \end{subfigure}

  \begin{subfigure}[b]{0.75\linewidth}
    \centering
    \includegraphics[width=\linewidth]{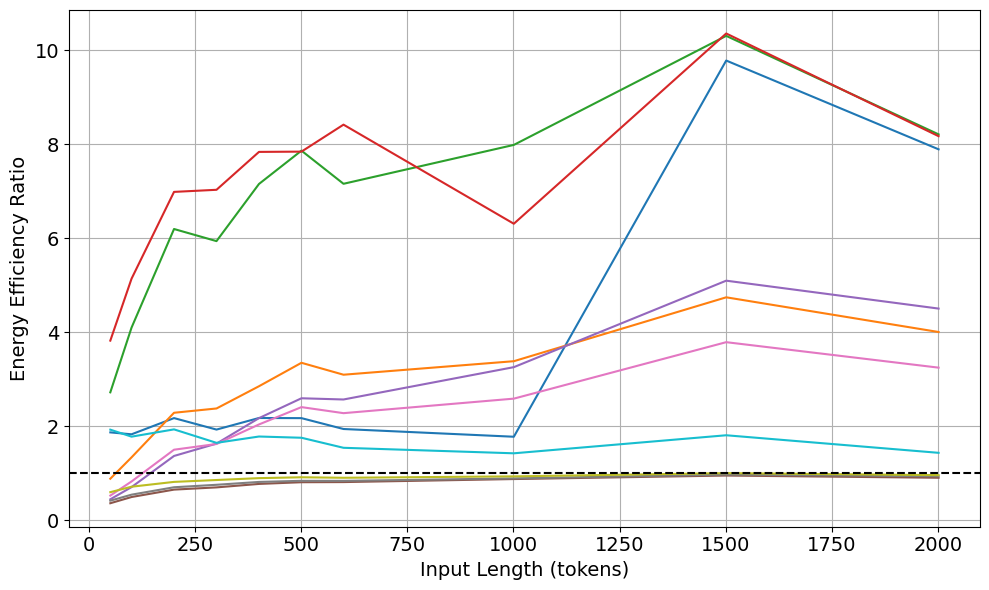}
    \caption{A100}
  \end{subfigure}

  \includegraphics[width=\linewidth, trim=0 105 0 65, clip]{Images/Quantization_Strategy_Legend_Flat.png}

  \caption{Energy Efficiency Ratio for Single-Token Decoding for Llama 2 7B Across Input Lengths on A6000 \& A100}
  \label{fig:energy_vs_input_length_combined}
  \vspace{-0.4cm}
\end{figure}

Figure~\ref{fig:energy_vs_input_length_combined} displays the energy curves of the inverse experiment, fixing output token lengths to 1 and varying input lengths. The results provide further evidence that energy savings increase steadily as input token lengths grow. Notably, HQQ and Quanto strategies, which rely on PyTorch's built-in matrix multiplication kernels, exhibit higher overhead with small input sizes, leading to relatively poor energy efficiency. However, as input length increases, the relative impact of this overhead diminishes, and their energy usage asymptotically approaches that of the vanilla FP32 baseline.

\begin{figure}[htbp]
\vspace{-1em}
\centering
\begin{subfigure}{0.8\linewidth}
  \includegraphics[width=\linewidth]{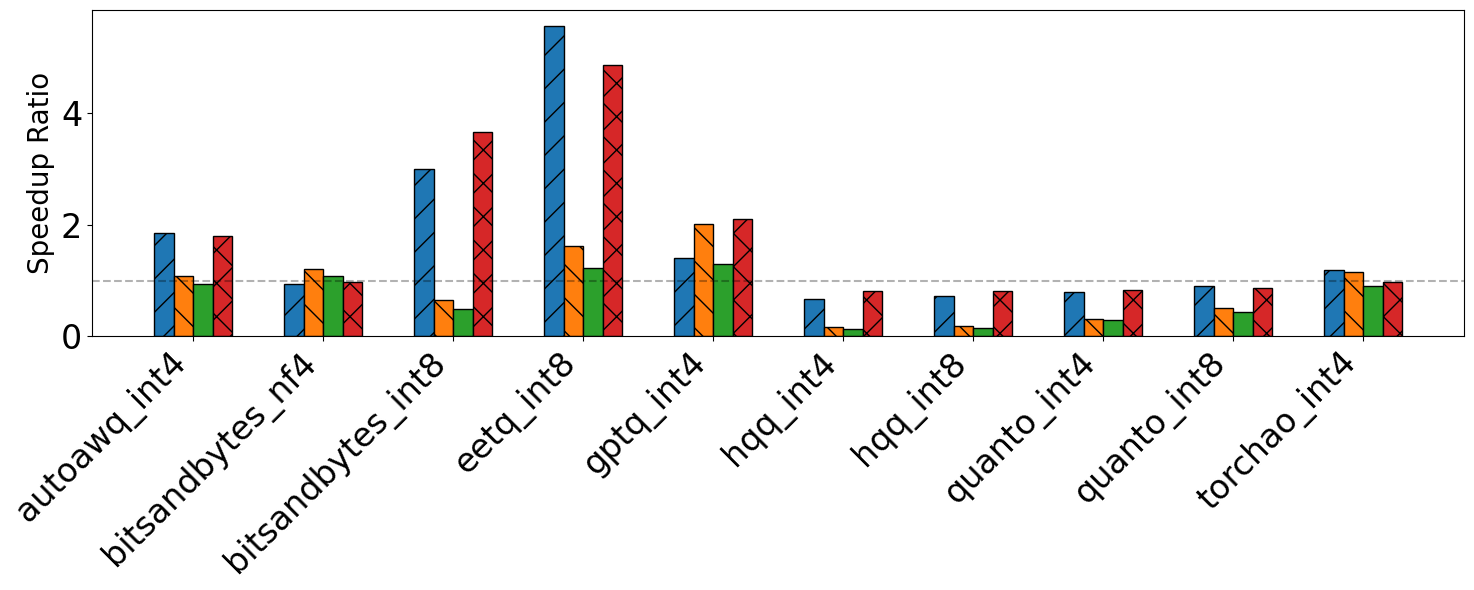}
  \caption{Speedup Ratio for Llama 2 7B}
\end{subfigure}
\begin{subfigure}{0.8\linewidth}
  \includegraphics[width=\linewidth]{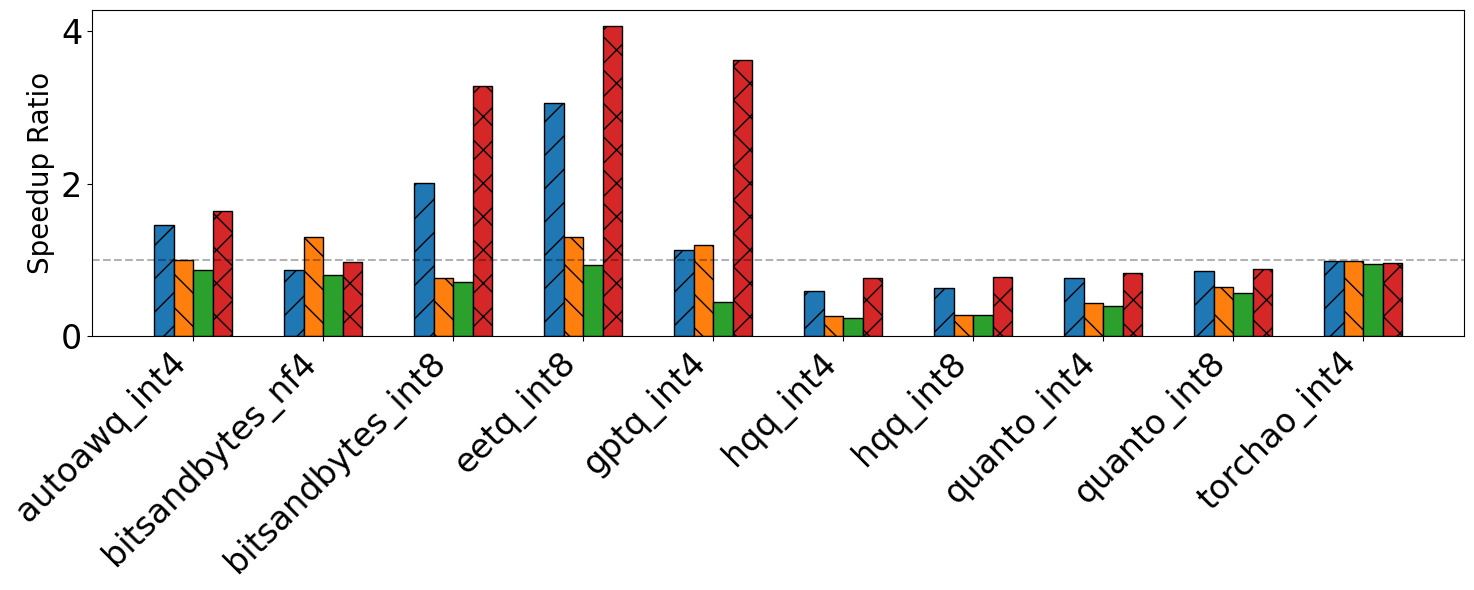}
  \caption{Speedup Ratio for Phi 3.5 Mini Instruct}
\end{subfigure}
\begin{subfigure}{0.8\linewidth}
  \includegraphics[width=\linewidth]{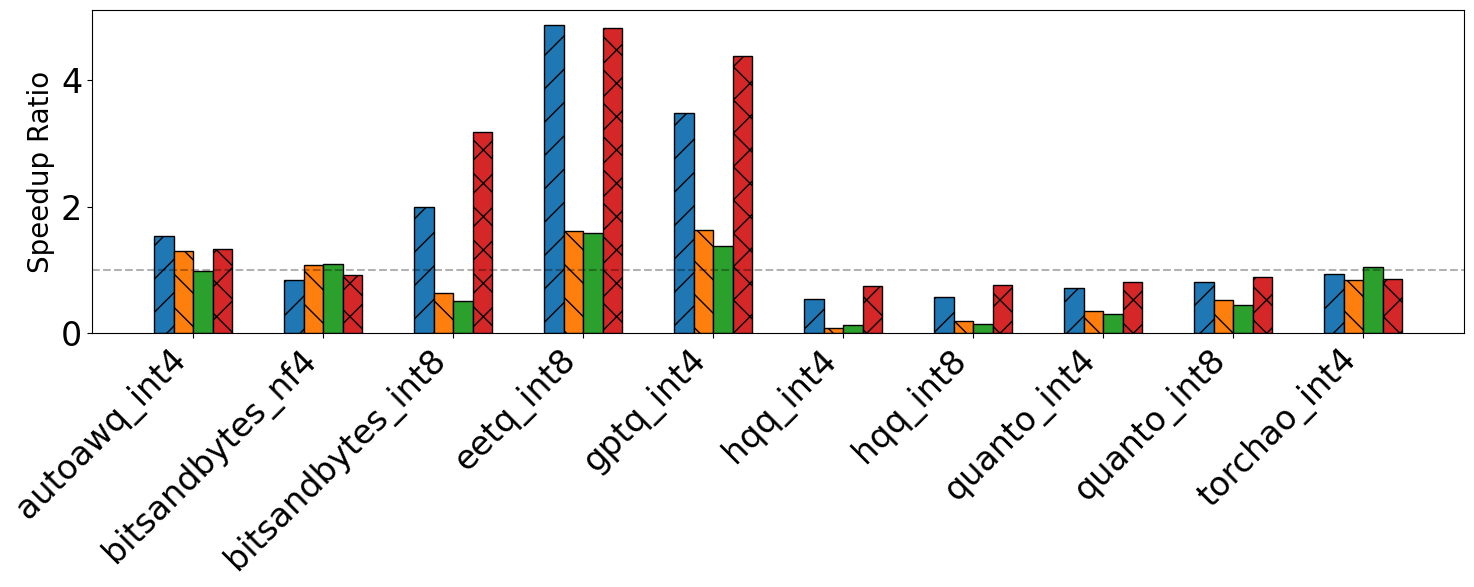}
  \caption{Speedup Ratio for Qwen 2.5 7B}
\end{subfigure}
\includegraphics[width=0.8\linewidth, trim=0 20 0 20, clip]{Images/Benchmark_Legend_Flat.png}

\caption{Speedup Ratio Across Models and Datasets on A6000}
\label{fig:speedupsA6000}
\vspace{-0.2cm}
\end{figure}

Figure~\ref{fig:speedupsA6000} displays the speedup ratios (i.e., baseline inference time divided by quantized time) for the same models, GPU, and quantization strategies as the original benchmarking experiments. The int8 methods again lead, with both EETQ int8 and Bitsandbytes int8 achieving the greatest speedups, confirming that their energy efficiency translates into faster inference throughput. The static 4-bit methods (GPTQ, AWQ, HQQ) and general quantization toolkits (Quanto, TorchAO) register more moderate speedups, often actually underperforming the FP32 baseline. Again, in Figure~\ref{fig:speedupsH100}, run on a newer H100 GPU and a wider subset of models, we can see speedups for supported quantization strategies. Interestingly, these speedups are slightly diminished on the newer hardware, even though the energy reduction is not.

This alignment between energy efficiency and speedup rankings suggests that, while quantization generally leads to energy savings, the corresponding reductions in inference time are not always as pronounced. Some quantization strategies reduce power/energy consumption more than they improve raw speed, highlighting the nuanced impact of low-bit precision on performance versus power.

Moreover, both energy efficiency and speedup are influenced by task characteristics, particularly input/output length. Quantization yields greater benefits for tasks involving short outputs, such as GLUE MNLI and MMLU, where the model only needs to generate a small number of tokens. These short-response workloads benefit significantly from reduced per-token computation costs. In contrast, tasks that require longer outputs, such as GSM8K and HumanEval, show only moderate improvements. Longer sequences entail attending to an expanding context and performing memory-intensive operations, which diminishes the relative advantage of quantization.

In conclusion, our findings empirically validate the thesis that quantization can significantly improve energy efficiency and often inference speed, but not in all cases. Sequence length plays a critical role in moderating these benefits. For tasks with short outputs, quantization consistently reduces both energy usage and latency, while its impact diminishes for tasks with long sequences. Thus, to fully capitalize on quantization, one must consider the context in which it is applied, as certain quantization methods yield diminishing returns or even negative speedups.

\begin{figure}[htbp]
\vspace{-1em}
\centering
\begin{subfigure}{0.8\linewidth}
  \includegraphics[width=\linewidth]{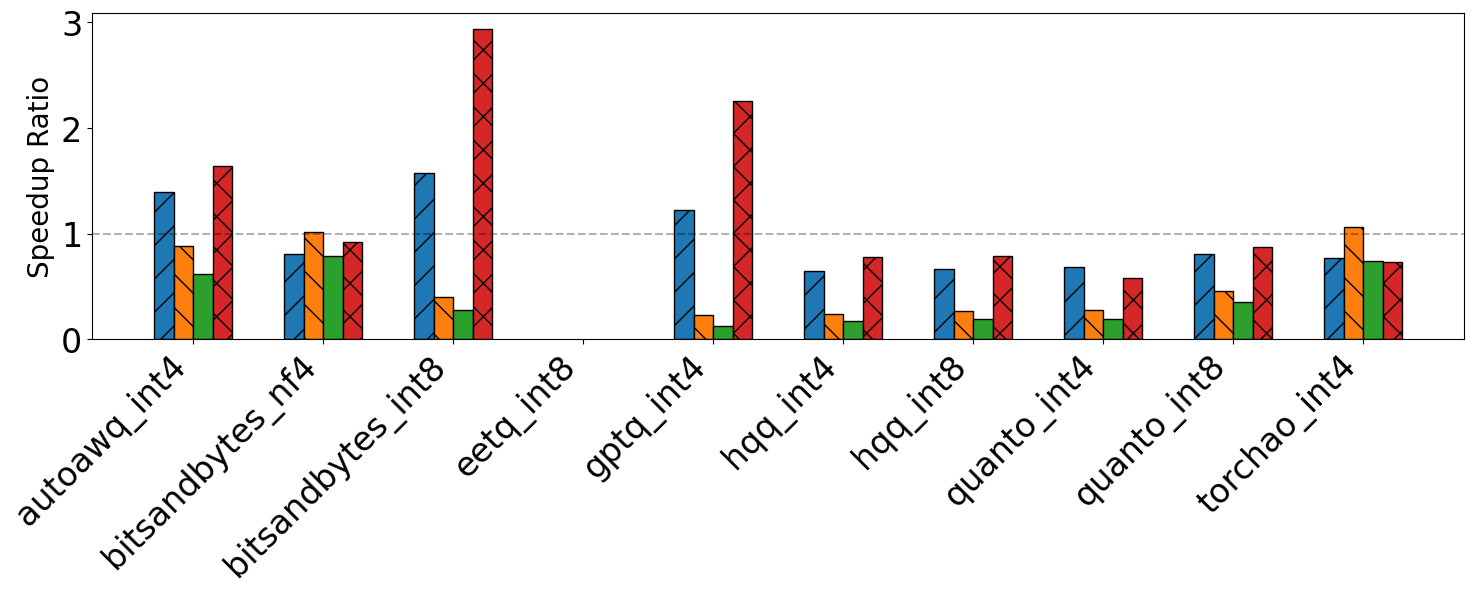}
  \caption{Speedup Ratio for Llama 2 7B}
\end{subfigure}
\begin{subfigure}{0.8\linewidth}
  \includegraphics[width=\linewidth]{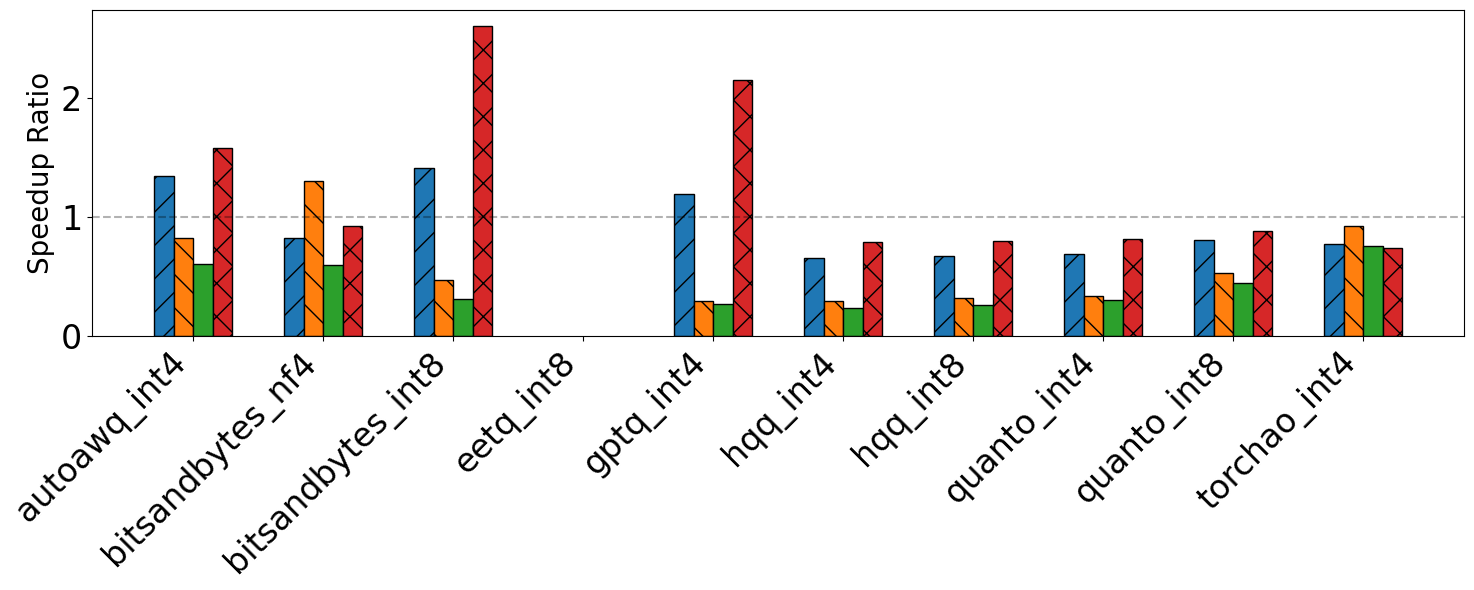}
  \caption{Speedup Ratio for Phi 3.5 Mini Instruct}
\end{subfigure}
\begin{subfigure}{0.8\linewidth}
  \includegraphics[width=\linewidth]{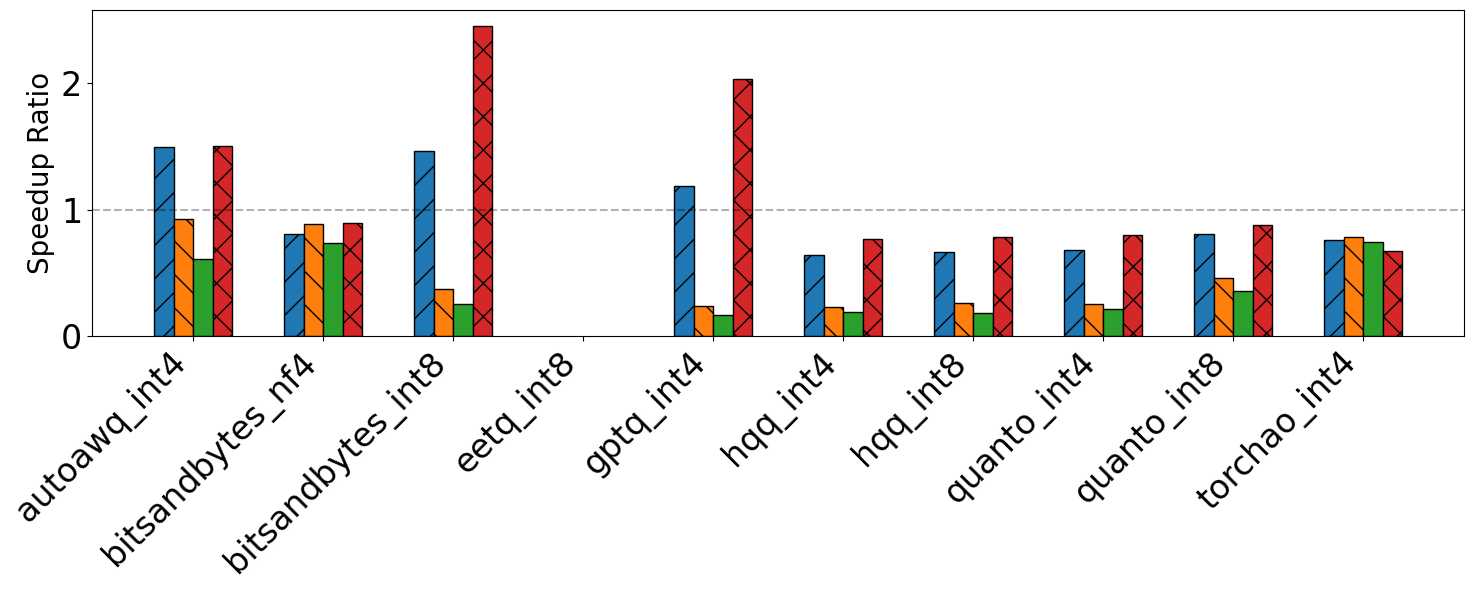}
  \caption{Speedup Ratio for Qwen 2.5 7B}
\end{subfigure}

\vspace{-0.2cm}
\end{figure}

\begin{figure}[htbp]\ContinuedFloat
\centering
\begin{subfigure}{0.8\linewidth}
  \includegraphics[width=\linewidth]{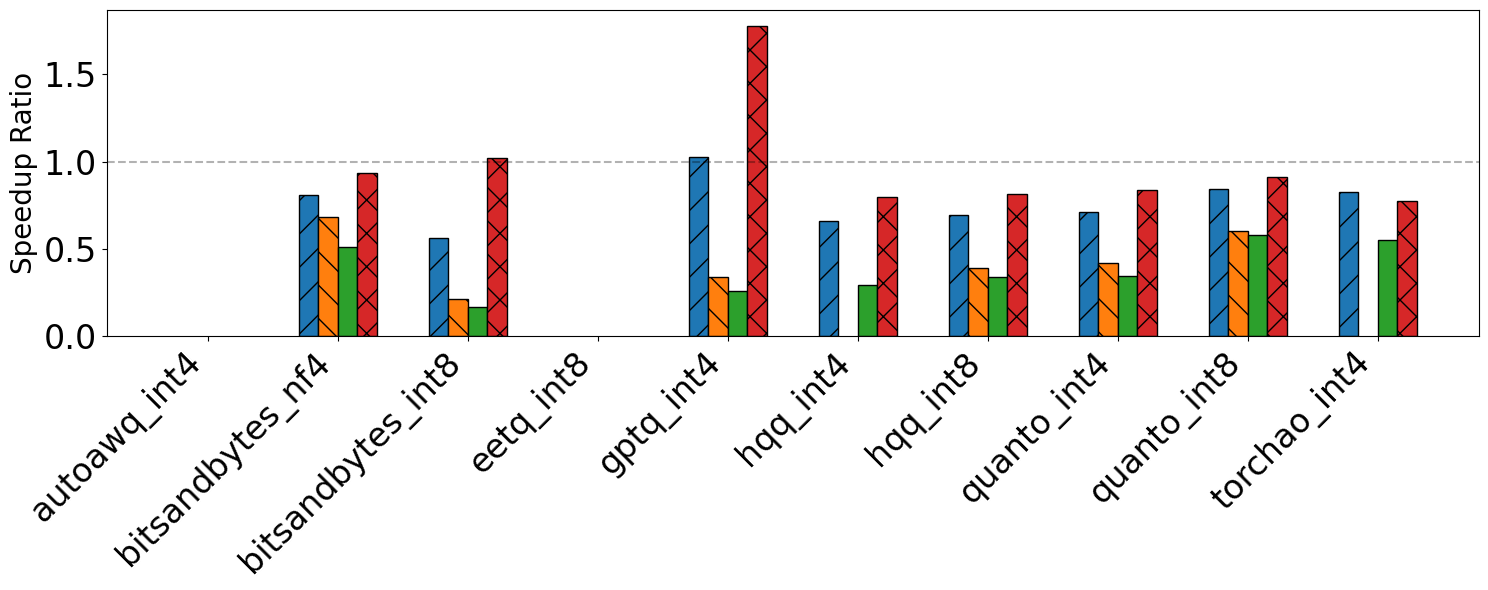}
  \caption{Speedup Ratio for OLMo 1B}
\end{subfigure}
\begin{subfigure}{0.8\linewidth}
  \includegraphics[width=\linewidth]{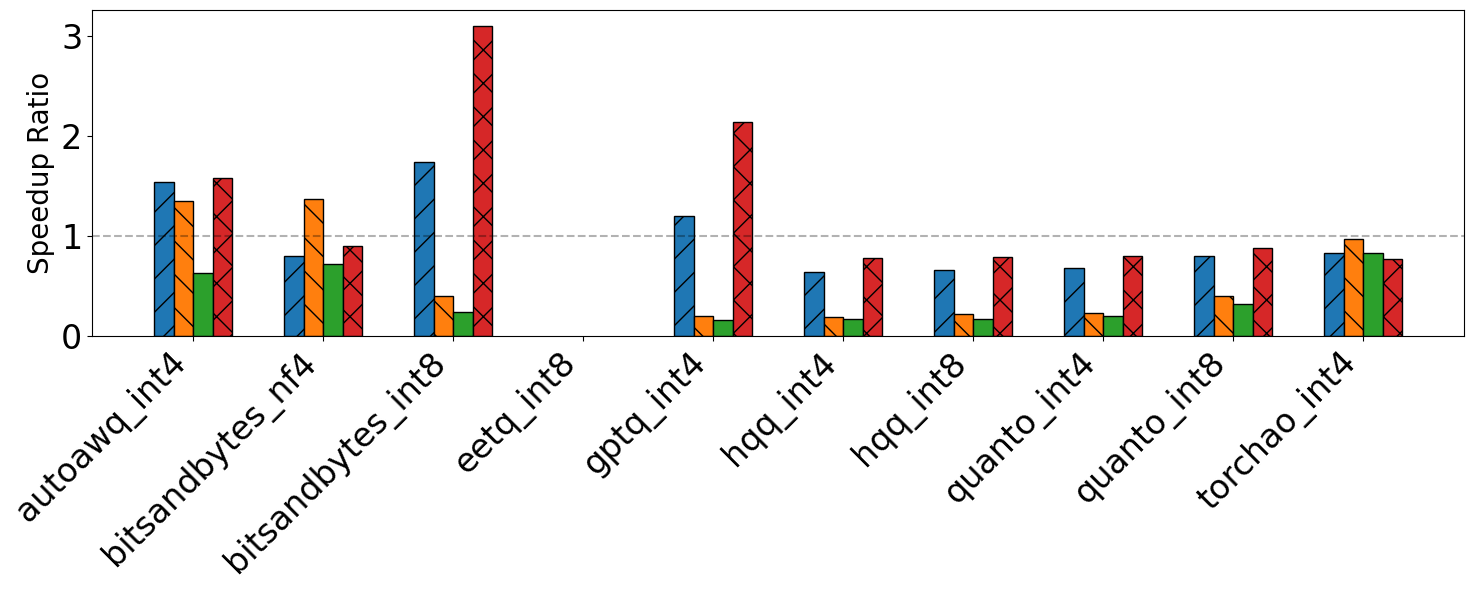}
  \caption{Speedup Ratio for Qwen 2.5 14B}
\end{subfigure}

\includegraphics[width=0.8\linewidth, trim=0 0 0 0, clip]{Images/Benchmark_Legend_Flat.png}
\caption{Speedup Ratio Across Models and Datasets on H100}
\label{fig:speedupsH100}
\end{figure}

\begin{center}
\vspace{0.5em}
\begin{tcolorbox}[
    colback=green!10!white, 
    colframe=green!50!black, 
    boxrule=0.5pt, 
    arc=2mm, 
    width=0.9\textwidth,
    box align=center
]

{\textbf{Takeaway 1:} Quantization strategies tend to deliver maximum energy efficiency when the input sequence is sufficiently long and the output sequence is short.}
\end{tcolorbox}
\vspace{0.5em}
\end{center}

Among int8 quantization methods, only Bitsandbytes int8 and EETQ int8 deliver meaningful energy savings across tasks. In contrast, Quanto int8 and HQQ int8 consume more energy than the FP32 baseline, despite using lower-bit representations.
For int4 methods, GPTQ int4 achieves the most substantial energy gains, followed by Bitsandbytes nf4 and TorchAO int4, and AutoAWQ int4 also demonstrating consistent improvements over FP32. However, HQQ int4 and Quanto int4 consume more energy than their FP32 counterparts, reflecting similar inefficiencies seen in int8 variants.

This divergence highlights that not all quantization methods, even within the same bitwidth, are equally energy-efficient: backend kernel optimizations, memory access patterns, and compute path efficiencies play a critical role in determining whether quantization yields real-world energy gains.

\begin{figure}[htbp]
  \centering
  \begin{subfigure}[b]{0.75\linewidth}
    \centering
    \includegraphics[width=\linewidth]{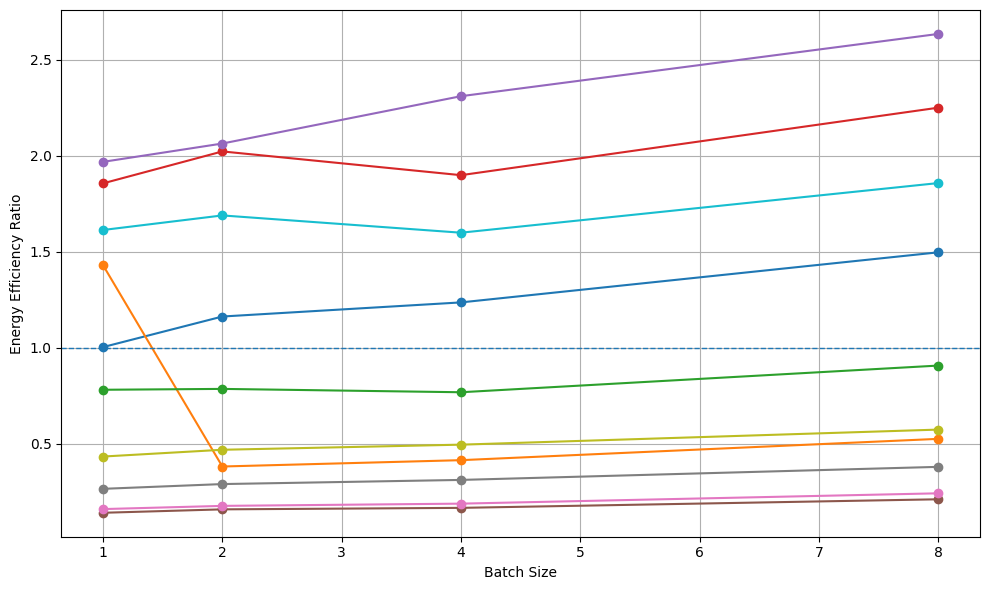}
    \caption{Input Length: 1, Output Length: 1000}
    \label{fig:energy_vs_batch_size_input_1_output_1000}
  \end{subfigure}

  \begin{subfigure}[b]{0.75\linewidth}
    \centering
    \includegraphics[width=\linewidth]{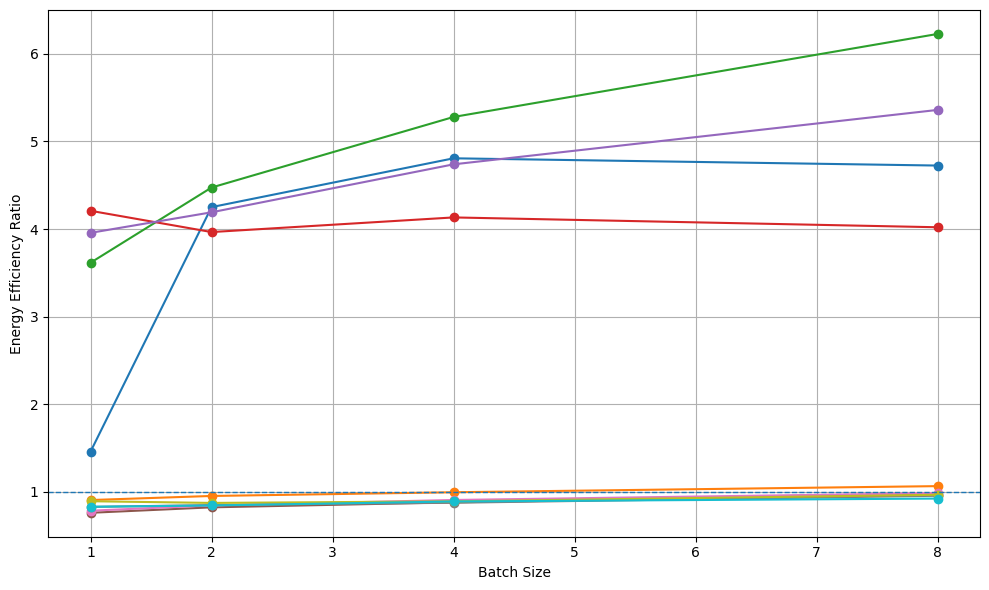}
    \caption{Input Length: 1000, Output Length: 1}
    \label{fig:energy_vs_batch_size_input_1000_output_1}
  \end{subfigure}

    \includegraphics[width=\linewidth, trim=0 105 0 65, clip]{Images/Quantization_Strategy_Legend_Flat.png}

  \caption{Energy Efficiency Ratio vs. Batch Size for Llama 2 7B with Varying Input and Output Lengths on A6000}
  \label{fig:batch_sizes}
  \vspace{-0.4cm}
\end{figure}

Figure~\ref{fig:energy_vs_batch_size_input_1_output_1000} plots per-sequence energy versus batch size for a 1-token prompt and 1000-token generation, a decode-dominated workload. Note that batch sizes start at 1 and follow powers of 2 until memory capacity errors arise with the FP32 model after batch size 8. The results show that energy efficiency improves for almost every quantization strategy as batch grows, but only by a small-medium margin.

\begin{itemize}
    \item \textbf{Int8 strategies (eetq int8, Bitsandbytes int8, hqq int8, quanto int8):} EETQ int8 has the highest energy efficiency ratios of the 8-bit group at every batch size. HQQ int8 starts far lower and stays several times below the leaders even after batching. Quanto int8 begins low but approaches the efficient range by batch 8.

    \item \textbf{Int4 strategies (autoawq int4, gptq int4, torchao int4, hqq int4, quanto int4, Bitsandbytes nf4):} GPTQ int4 traces the highest curves overall, with TorchAO int 4 slightly behind, and AutoAWQ int4 lagging even further. Bitsandbytes nf4 displays the only fluctuating results in the experiment at batch size 1, but stabilizes below the vanilla FP32 baseline for higher batch sizes. HQQ int4 and Quanto int4 are worst performing overall, barely approaching 0.25x energy efficiency.
\end{itemize}

Figure~\ref{fig:energy_vs_batch_size_input_1000_output_1} shows per-sequence energy for a 1000-token input and 1-token output, a prefill-dominated workload where large attention matrix multiplications dominate compute.

\begin{itemize}
    \item \textbf{Upper group (AutoAWQ int4, EETQ int8, Bitsandbytes int8, GPTQ int4:} Bitsandbytes int8 is the single highest curve, about 3.6x at batch size 1 and peaking near 6.2x at batch size 8. GPTQ int4 and AutoAWQ int4 also follow a similar trend where energy efficiency increases as batch size increases; only EETQ int8 seems relative flat at about a 4x increase regardless of batch size.

    \item \textbf{Lower group (HQQ int4/int8, Bitsandbytes nf4, TorchAO int4, Quanto int4/int8:} Bitsandbytes nf4 is the only strategy that falls in the lower group that outperforms vanilla FP32 for batch size 8. Quanto int4/int8 sit relatively flat regardless of batch size. It is interesting that these never break the asymptote of vanilla FP32 performance, showing us that the underlying matmuls are not any more efficient and that the kernel launch overhead is substantial.
 
\end{itemize}

For long-context inference, efficient int8 and selected 4-bit kernels already deliver large savings at batch 1. Batching to 4–8 captures most remaining benefit; larger batches show little or negative marginal improvement once the GPU memory subsystem is saturated.

\begin{figure}[htbp]
  \centering
  \begin{subfigure}[b]{0.75\linewidth}
    \centering
    \includegraphics[width=\linewidth]{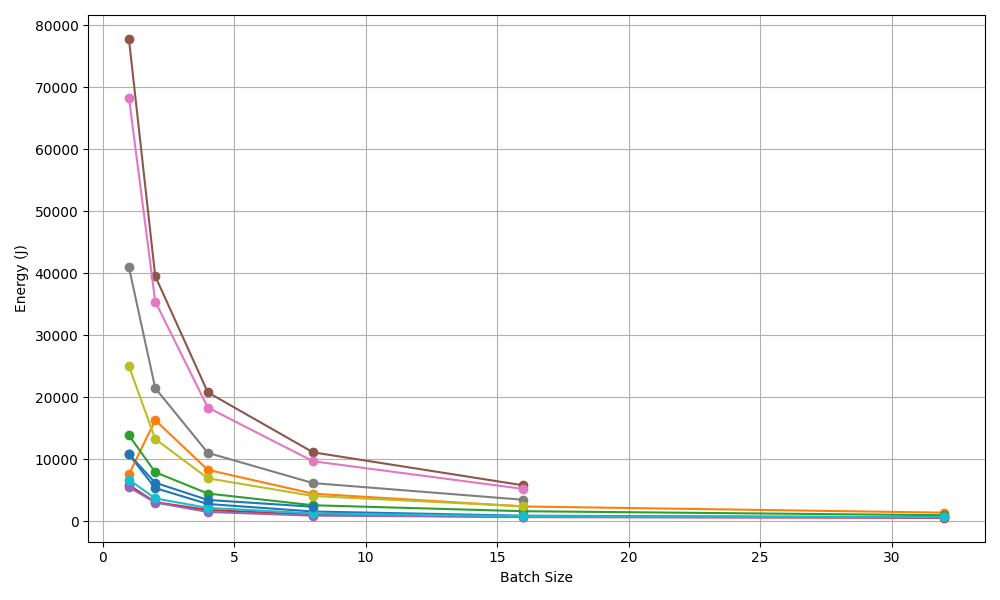}
    \caption{Input Length: 1, Output Length: 1000}
    \label{fig:energy_vs_batch_size_input_1_output_1000_absolute}
  \end{subfigure}

  \begin{subfigure}[b]{0.75\linewidth}
    \centering
    \includegraphics[width=\linewidth]{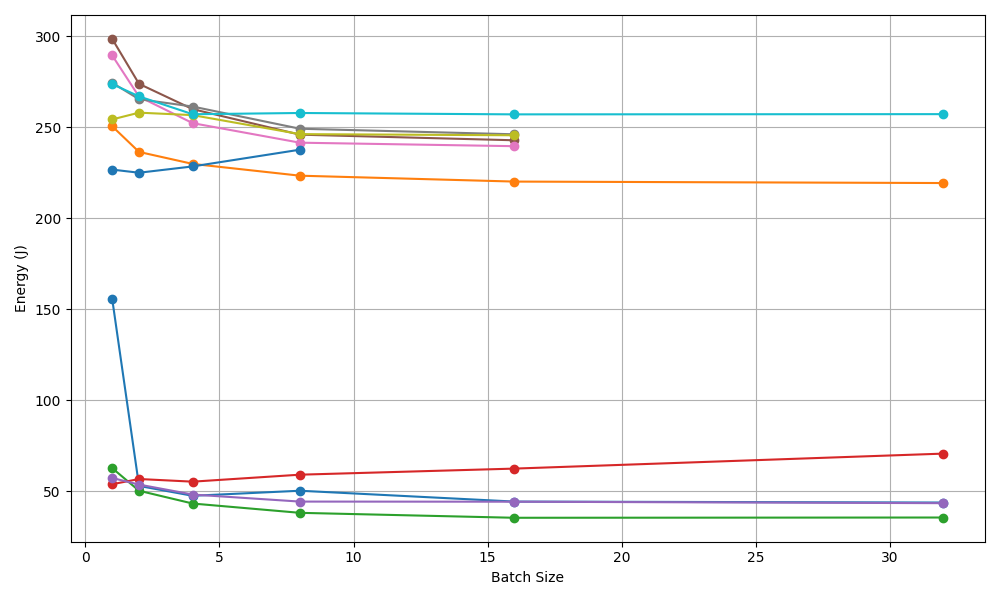}
    \caption{Input Length: 1000, Output Length: 1}
    \label{fig:energy_vs_batch_size_input_1000_output_1_absolute}
  \end{subfigure}

    \includegraphics[width=\linewidth, trim=0 105 0 65, clip]{Images/Quantization_Strategy_Legend_Flat.png}

  \caption{Absolute Energy Usage vs. Batch Size for Llama 2 7B with Varying Input and Output Lengths on A6000}
  \label{fig:batch_sizes_absolute}
  \vspace{-0.4cm}
\end{figure}

Looking at absolute energy usage in Figure~\ref{fig:batch_sizes_absolute}, we can see that for short-input, long-output tasks, energy consumption goes down rather steeply as batch sizes increase. Conversely it seems to stay relatively flat when batch sizes increase for long-input, short output tasks. This tells us that increasing batch size has a larger effect on the decode phase, more efficiently parallelizing sequential decode operations than large prefill matmuls. This is due to memory bottlenecks with large input matrices and already efficient kernel implementations.

\begin{center}
\begin{tcolorbox}[
    colback=green!10!white, 
    colframe=green!50!black, 
    boxrule=0.5pt, 
    arc=2mm, 
    width=0.9\textwidth,
    box align=center
]

{\textbf{Takeaway 2:} Increasing batch size for quantized models causes gradual but minimal improvements in energy efficiency relative to full-precision baselines} 
\end{tcolorbox}
\end{center}

\section{Accuracy Degradation}
\label{sec:accuracy_degradation}

While quantization provides substantial memory savings, it inherently risks degrading model accuracy. Table~\ref{tab:quantized_model_results} presents the performance of each quantized model across all evaluated benchmarks, with cells colored according to the degree of degradation in performance relative to the vanilla FP32 baseline. Cells are colored red if performance drops by more than 5\%, and colored yellow if performance drops between 2.5\% and 5\% compared to the vanilla FP32 model.

\begin{table*}[htbp]
\centering
\scriptsize
\resizebox{\textwidth}{!}{%
\begin{tabular}{llccccccccccc}
\toprule
\textbf{Model} & \textbf{Benchmark} & \textbf{vanilla FP32} & \textbf{autoawq int4} & \textbf{Bitsandbytes nf4} & \textbf{Bitsandbytes int8} & \textbf{eetq int8} & \textbf{gptq int4} & \textbf{hqq int4} & \textbf{hqq int8} & \textbf{quanto int4} & \textbf{quanto int8} & \textbf{torchao int4} \\
\midrule
\multirow{4}{*}{Llama 2 7B} & GLUE MNLI & 0.4318 & 0.4285 & 0.4235 & 0.4290 & 0.4304 & 0.4268 & 0.4325 & 0.4320 & \cellcolor{red!25}{0.3696} & 0.4316 & \cellcolor{yellow!25}{0.4138} \\
& GSM8K (10\%) & 0.1603 & \cellcolor{red!25}{0.1298} & \cellcolor{yellow!25}{0.1527} & 0.1832 & 0.1679 & \cellcolor{red!25}{0.0916} & \cellcolor{red!25}{0.1221} & \cellcolor{yellow!25}{0.1527} & \cellcolor{yellow!25}{0.1527} & \cellcolor{yellow!25}{0.1527} & \cellcolor{red!25}{0.0992} \\
& HumanEval & 0.1159 & 0.1280 & \cellcolor{red!25}{0.1098} & 0.1280 & 0.1220 & 0.1280 & 0.1280 & 0.1159 & 0.1280 & 0.1280 & 0.1341 \\
& MMLU & 0.4469 & 0.4445 & \cellcolor{red!25}{0.4153} & 0.4415 & 0.4449 & \cellcolor{yellow!25}{0.4271} & \cellcolor{yellow!25}{0.4356} & 0.4440 & \cellcolor{yellow!25}{0.4300} & 0.4468 & \cellcolor{red!25}{0.4242} \\
\multirow{4}{*}{Phi 3.5 Mini Instruct} & GLUE MNLI & 0.7465 & \cellcolor{yellow!25}{0.7152} & \cellcolor{red!25}{0.6969} & \cellcolor{yellow!25}{0.7237} & 0.7413 & \cellcolor{red!25}{0.6803} & \cellcolor{red!25}{0.6630} & 0.7477 & \cellcolor{red!25}{0.6916} & 0.7457 & \cellcolor{red!25}{0.6694} \\
& GSM8K (10\%) & 0.7786 & 0.7634 & \cellcolor{red!25}{0.6489} & 0.8092 & 0.7634 & \cellcolor{red!25}{0.7252} & \cellcolor{red!25}{0.6718} & 0.7786 & 0.7710 & 0.7939 & \cellcolor{yellow!25}{0.7481} \\
& HumanEval & 0.5183 & \cellcolor{red!25}{0.4329} & \cellcolor{red!25}{0.1707} & 0.5183 & 0.5183 & \cellcolor{red!25}{0.4451} & \cellcolor{red!25}{0.4268} & 0.5305 & \cellcolor{red!25}{0.2866} & 0.5061 & \cellcolor{red!25}{0.3171} \\
& MMLU & 0.6807 & \cellcolor{yellow!25}{0.6592} & \cellcolor{yellow!25}{0.6480} & 0.6679 & 0.6803 & \cellcolor{yellow!25}{0.6491} & \cellcolor{red!25}{0.6033} & 0.6781 & \cellcolor{red!25}{0.6466} & 0.6813 & \cellcolor{red!25}{0.6389} \\
\multirow{4}{*}{Qwen 2.5 7B} & GLUE MNLI & 0.8448 & \cellcolor{yellow!25}{0.8161} & \cellcolor{red!25}{0.7769} & 0.8425 & 0.8426 & \cellcolor{yellow!25}{0.8115} & 0.8259 & 0.8452 & \cellcolor{red!25}{0.7944} & 0.8453 & 0.8355 \\
& GSM8K (10\%) & 0.7405 & \cellcolor{red!25}{0.7023} & \cellcolor{red!25}{0.6336} & 0.7557 & \cellcolor{yellow!25}{0.7176} & 0.8244 & 0.7634 & 0.7481 & 0.7252 & 0.7634 & \cellcolor{red!25}{0.6718} \\
& HumanEval & 0.3963 & 0.4146 & 0.4085 & 0.4329 & 0.4024 & \cellcolor{red!25}{0.3598} & \cellcolor{yellow!25}{0.3780} & 0.4024 & 0.4268 & 0.4024 & \cellcolor{red!25}{0.3720} \\
& MMLU & 0.7173 & 0.7123 & \cellcolor{yellow!25}{0.6991} & 0.7171 & 0.7169 & 0.7044 & 0.7078 & 0.7163 & 0.7060 & 0.7178 & 0.7069 \\
\bottomrule
\end{tabular}%
}
\caption{Quantized Model Accuracies Across Benchmarks}
\label{tab:quantized_model_results}
\vspace{-0.5cm}
\end{table*}

Overall, we observe that 8-bit quantization strategies, particularly Bitsandbytes int8, EETQ int8, HQQ int8, and Quanto int8, consistently preserve model performance across diverse tasks, maintaining accuracies within 2.5\% of the full-precision baseline in the majority of cases. This demonstrates that modern int8 methods are sufficiently precise for practical deployment, maintaining high accuracy across tasks. Relating back to energy savings, only Bitsandbytes int8 and EETQ int8 yielded meaningful reductions in energy usage across tasks, whereas Quanto int8 and HQQ int8 actually consumed more energy than the FP32 baselines.

\begin{center}
\begin{tcolorbox}[
    colback=green!10!white, 
    colframe=green!50!black, 
    boxrule=0.5pt, 
    arc=2mm, 
    width=0.9\textwidth ,
]
\textbf{Takeaway 3:} From strategies that display negligible accuracy degradation, EETQ int8 offers the highest energy savings, up to $4\times$ compared to FP32 on short text generation. Bitsandbytes int8 is a strong alternative.
\end{tcolorbox}
\end{center}

In contrast, 4-bit quantization introduces greater variability, with accuracy losses highly dependent on both the quantization method and the task. Among 4-bit strategies, AutoAWQ int4 exhibits relatively stronger robustness, often maintaining degradation within 2.5\% or 5\%, whereas methods such as Bitsandbytes nf4, GPTQ int4, HQQ int4, Quanto int4, and TorchAO int4 show substantial drops across the majority of tasks.

Finally, model size and training quality appear to mediate robustness. Larger, more capable models (e.g., Llama 2 7B, Qwen 2.5 7B) tend to withstand aggressive quantization more effectively than smaller models (e.g., Phi 3.5 Mini Instruct), particularly at 4-bit precision. This aligns with intuition: larger models are better able to absorb quantization noise due to greater redundancy in their parameter space. In over-parameterized networks, this redundancy provides a natural buffer against perturbations, enhancing robustness to precision loss.

\begin{center}
\begin{tcolorbox}[
    colback=green!10!white, 
    colframe=green!50!black, 
    boxrule=0.5pt, 
    arc=2mm, 
    width=0.9\textwidth ,
]
\textbf{Takeaway 4:} Larger models tend to tolerate low-bit quantization better than smaller ones due to their redundant parameter space naturally absorbing quantization noise.
\end{tcolorbox}
\end{center}

In summary, while 8-bit quantization offers a reliable path toward compression with minimal accuracy sacrifice, 4-bit strategies must be selected and applied carefully, with attention to model size, task type, and acceptable quality thresholds for the deployment scenario.

\section{Runtime as a Proxy for Energy Efficiency}

As shown in Figure~\ref{fig:energy_vs_time_mmlu}, there is a strong correlation between benchmark runtime and energy usage across quantization strategies. This trend is consistent across all three models and all four benchmarks, with Pearson correlation coefficients ranging from 0.989 to 1.000. On average, the correlation is extremely high at 0.9966, with a standard deviation of just 0.0033, indicating very little variation across tasks and models.

This exceptionally strong correlation implies that optimizing
for inference speed is an effective proxy for optimizing energy
efficiency. In practice, strategies that reduce runtime, such as low-bit quantization, fused kernel execution, or more efficient attention
mechanisms, will almost always result in proportional reductions in
total energy usage. Therefore, it can sometimes be simpler and more practical to focus on minimizing latency, which serves as a reliable method to indirectly reduce the environmental footprint of LLM inference.

\begin{figure}[htbp]
\centering

\begin{subfigure}{ 0.55\linewidth}
  \centering
  \includegraphics[width=\linewidth]{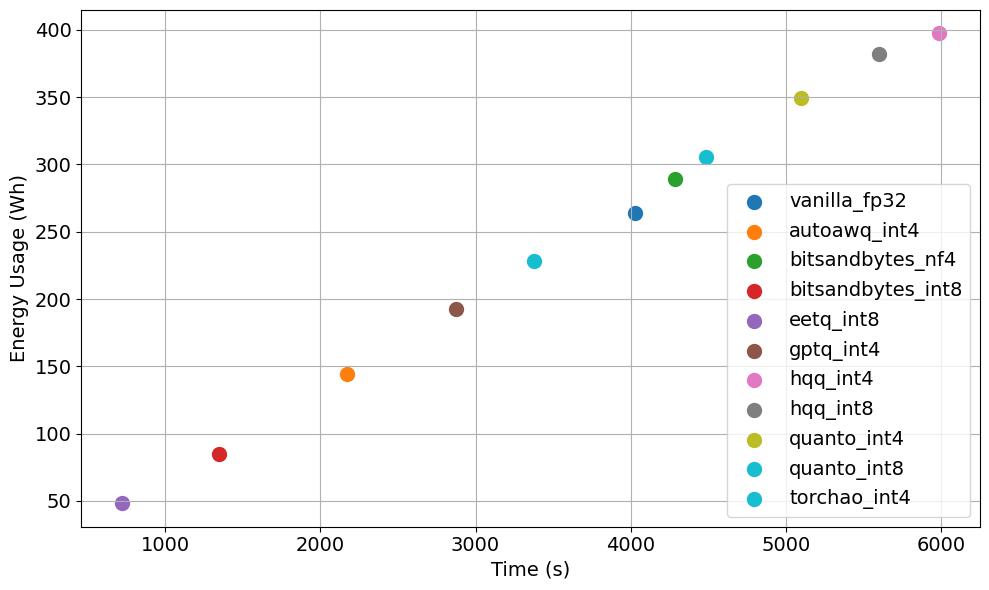}
  \caption{Llama 2 7B on GLUE MNLI}
\end{subfigure}\par\medskip

\begin{subfigure}{ 0.55\linewidth}
  \centering
  \includegraphics[width=\linewidth]{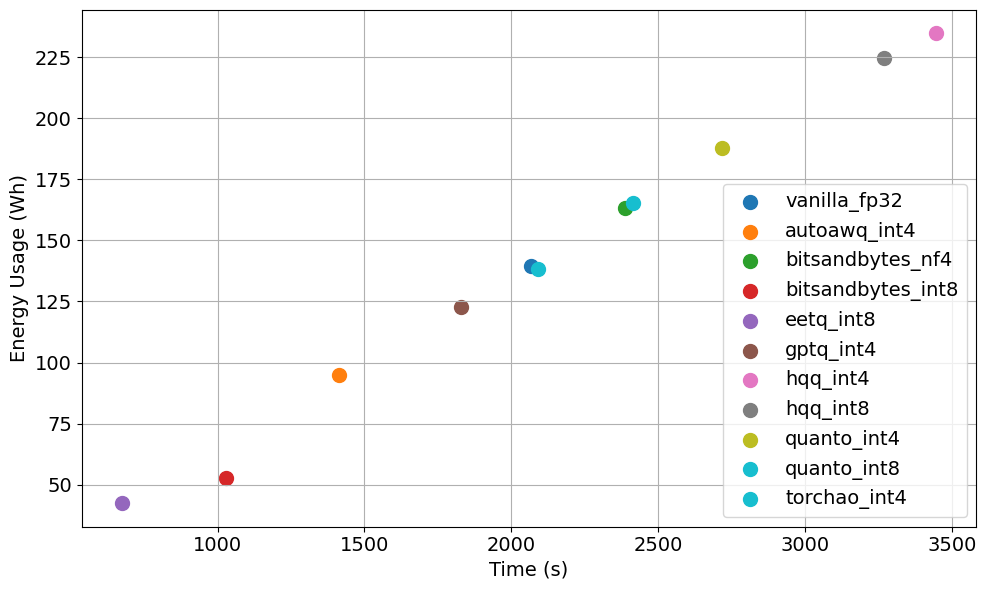}
  \caption{Phi 3.5 Mini Instruct on GLUE MNLI}
\end{subfigure}\par\medskip

\begin{subfigure}{ 0.55\linewidth}
  \centering
  \includegraphics[width=\linewidth]{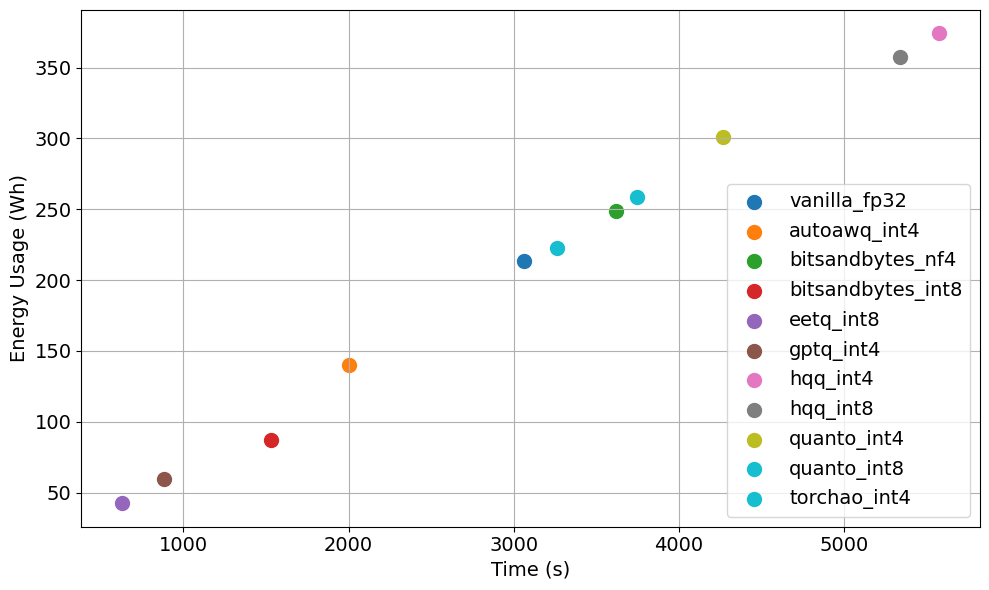}
  \caption{Qwen 2.5 7B on GLUE MNLI}
\end{subfigure}

\label{fig:energy_vs_time_glue}
\end{figure}

\begin{figure}[htbp]\ContinuedFloat
\centering

\begin{subfigure}{ 0.55\linewidth}
  \centering
  \includegraphics[width=\linewidth]{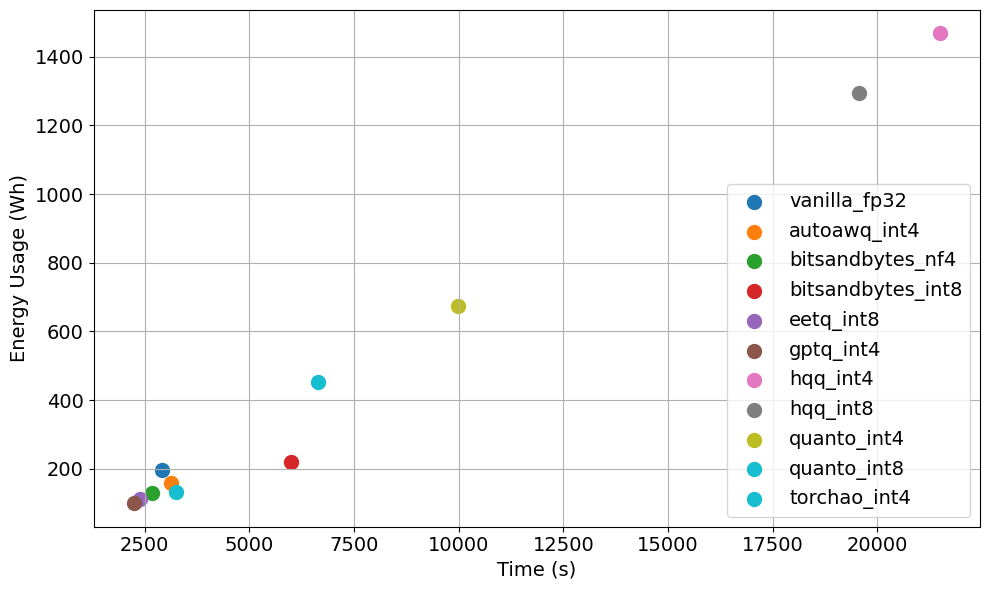}
  \caption{Llama 2 7B on HumanEval}
\end{subfigure}\par\medskip

\begin{subfigure}{ 0.55\linewidth}
  \centering
  \includegraphics[width=\linewidth]{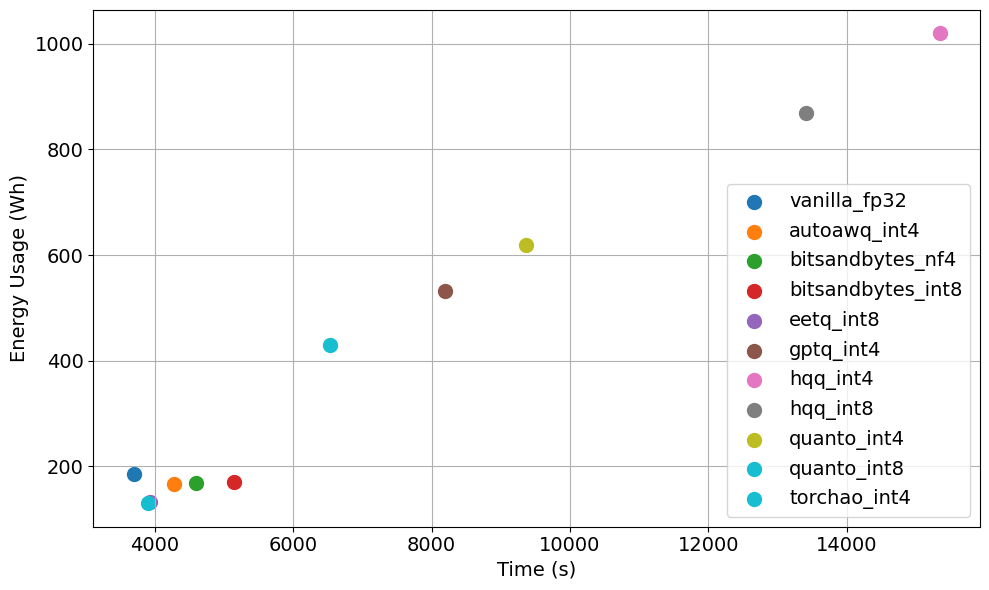}
  \caption{Phi 3.5 Mini Instruct on HumanEval}
\end{subfigure}\par\medskip

\begin{subfigure}{ 0.55\linewidth}
  \centering
  \includegraphics[width=\linewidth]{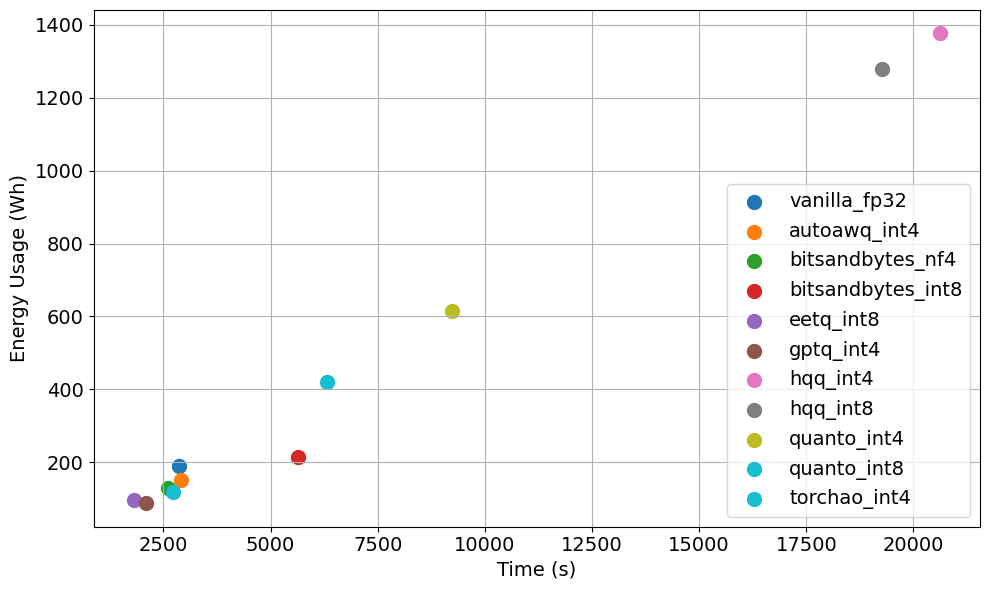}
  \caption{Qwen 2.5 7B on HumanEval}
\end{subfigure}

\label{fig:energy_vs_time_humaneval}
\end{figure}

\begin{figure}[htbp]\ContinuedFloat
\centering

\begin{subfigure}{ 0.55\linewidth}
  \centering
  \includegraphics[width=\linewidth]{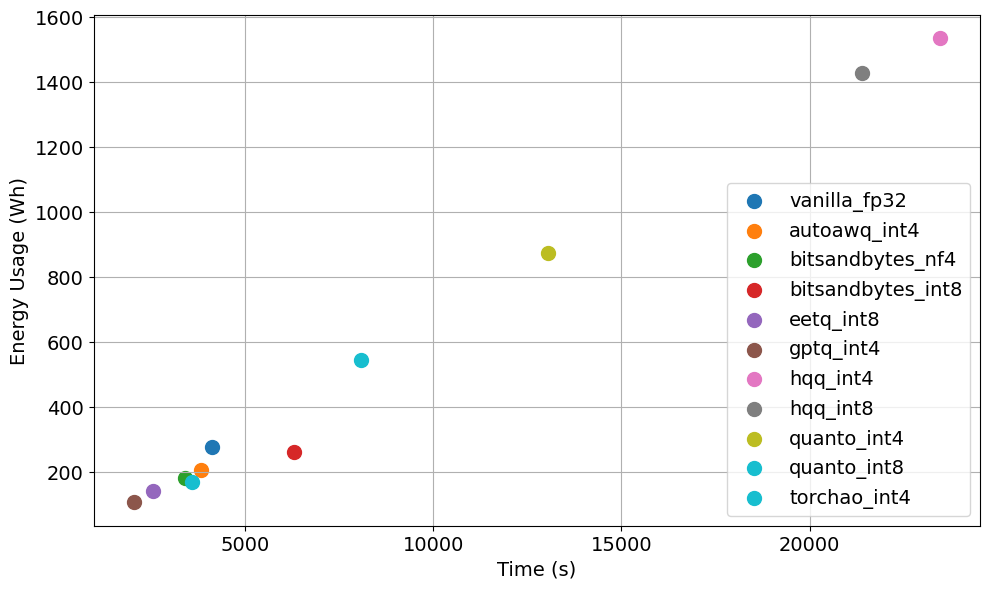}
  \caption{Llama 2 7B on GSM8K}
\end{subfigure}\par\medskip

\begin{subfigure}{ 0.55\linewidth}
  \centering
  \includegraphics[width=\linewidth]{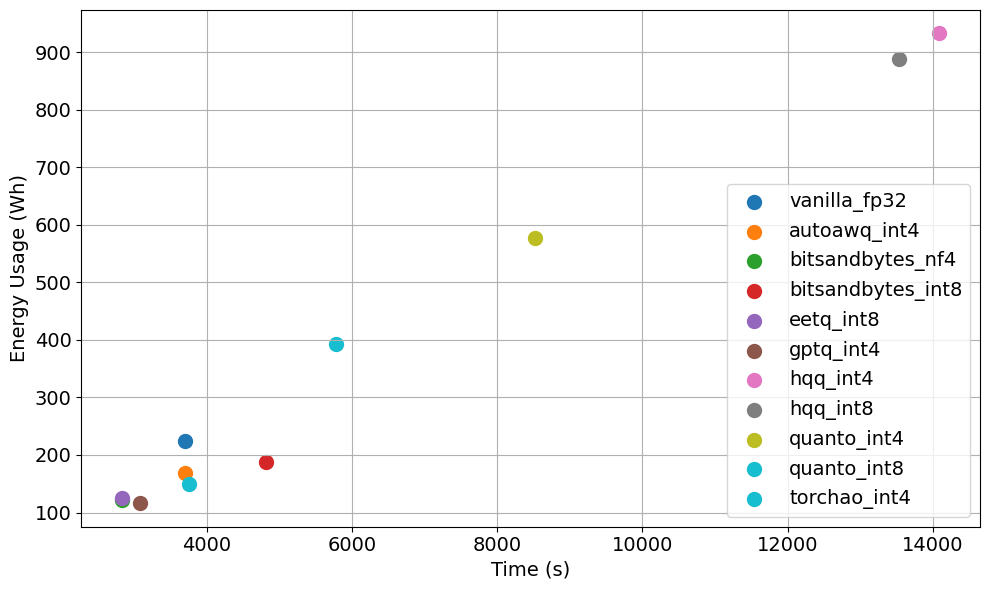}
  \caption{Phi 3.5 Mini Instruct on GSM8K}
\end{subfigure}\par\medskip

\begin{subfigure}{ 0.55\linewidth}
  \centering
  \includegraphics[width=\linewidth]{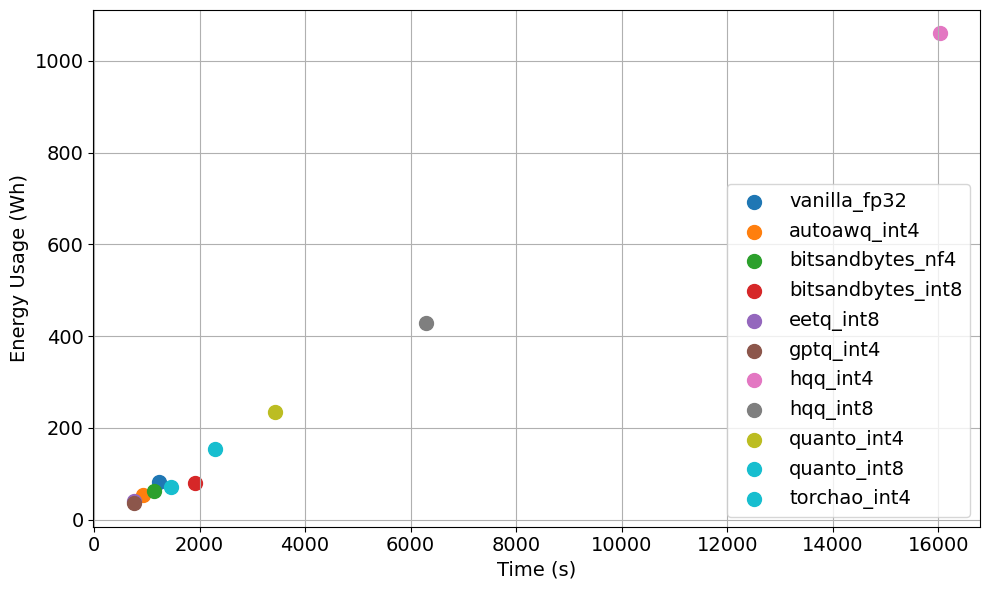}
  \caption{Qwen 2.5 7B on GSM8K}
\end{subfigure}

\label{fig:energy_vs_time_gsm8k}
\end{figure}

\begin{figure}[htbp]\ContinuedFloat
\centering

\begin{subfigure}{ 0.55\linewidth}
  \centering
  \includegraphics[width=\linewidth]{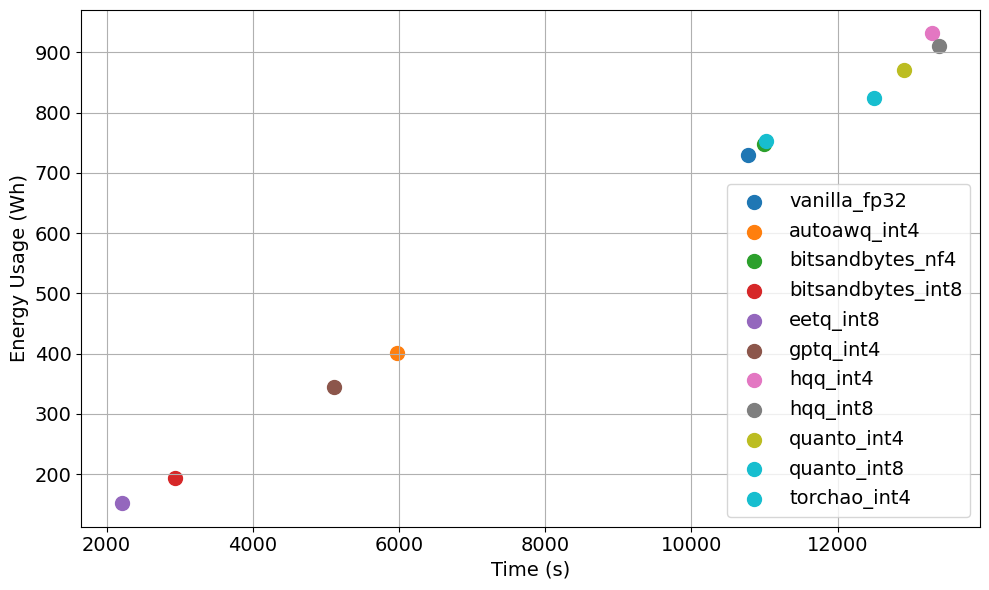}
  \caption{Llama 2 7B on MMLU}
\end{subfigure}\par\medskip

\begin{subfigure}{ 0.55\linewidth}
  \centering
  \includegraphics[width=\linewidth]{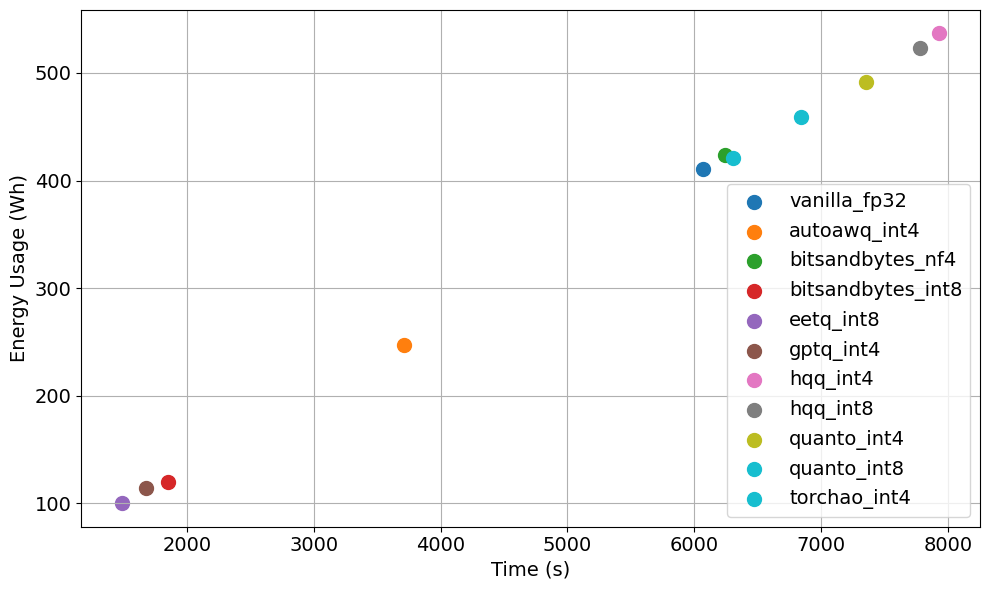}
  \caption{Phi 3.5 Mini Instruct on MMLU}
\end{subfigure}\par\medskip

\begin{subfigure}{ 0.55\linewidth}
  \centering
  \includegraphics[width=\linewidth]{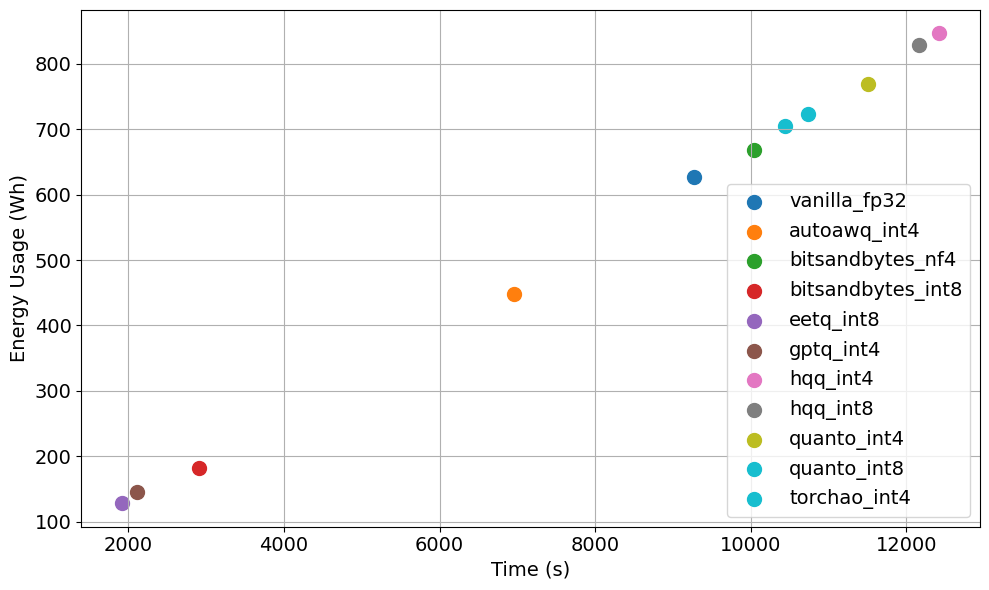}
  \caption{Qwen 2.5 7B on MMLU}
\end{subfigure}

\caption{Energy Usage vs Time on A6000}
\label{fig:energy_vs_time_mmlu}
\end{figure}

\begin{center}
\begin{tcolorbox}[
    colback=green!10!white, 
    colframe=green!50!black, 
    boxrule=0.5pt, 
    arc=2mm, 
    width=0.9\textwidth,
    box align=center
]
\textbf{Takeaway 5:} Reducing inference latency is an effective proxy to lower energy consumption, given the strong correlation between runtime and energy usage in practice.
\end{tcolorbox}
\end{center}

\section{Memory Usage}

Although memory reduction is the most well-established benefit of quantization, we briefly summarize the relative efficiency of each method to guide practitioners seeking trade-off decisions.

Table~\ref{tab:memory_efficiency_ratio} summarizes the average memory efficiency ratio for each quantization strategy, relative to the vanilla FP32 baseline, aggregated across all models. All strategies achieved notable memory savings, with 4-bit approaches delivering the most substantial gains. AutoAWQ int4 and TorchAO int4 were the most memory-efficient, achieving $5.73\times$ and $5.55\times$ improvements respectively. GPTQ int4, Quanto int4, and Bitsandbytes nf4 also demonstrated strong memory reductions, exceeding $4\times$ efficiency relative to FP32.

\begin{table}[htbp]
\centering
\footnotesize
\begin{tabular}{l c}
\hline
\textbf{Quantization Strategy} & \textbf{Avg Memory Efficiency Ratio} \\
\hline
autoawq int4 & 5.73x \\
Bitsandbytes nf4 & 5.11x \\
Bitsandbytes int8 & 3.53x \\
eetq int8 & 3.87x \\
gptq int4 & 4.57x \\
hqq int4 & 3.52x \\
hqq int8 & 2.52x \\
quanto int4 & 4.03x \\
quanto int8 & 2.87x \\
torchao int4 & 5.55x \\
\hline
\end{tabular}
\caption{Average Memory Efficiency Ratio Across Models}
\label{tab:memory_efficiency_ratio}
\end{table}

Among 8-bit methods, EETQ int8 provided a $3.87\times$ memory efficiency ratio, outperforming Bitsandbytes int8 ($3.53\times$) and Quanto int8 ($2.87\times$). HQQ int8 exhibited the lowest memory savings overall at $2.52\times$, reflecting trade-offs made to preserve higher model accuracy during weight quantization.

These results reinforce that memory-constrained environments, such as edge deployments or limited VRAM servers, stand to benefit substantially from quantization strategies, albeit with varying trade-offs in computational overhead and accuracy that were explored in earlier sections.

\begin{figure}[htbp]
  \centering
  \begin{subfigure}[b]{\linewidth}
    \centering
    \includegraphics[width=0.8\linewidth]{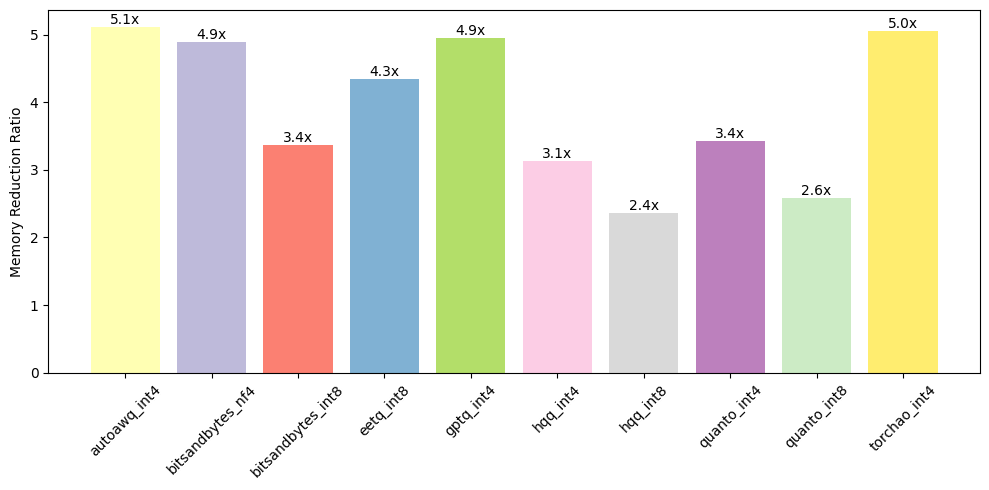}
    \caption{Llama 2 7B}
    \label{fig:memory_llama}
  \end{subfigure}

  \begin{subfigure}[b]{\linewidth}
    \centering
    \includegraphics[width=0.8\linewidth]{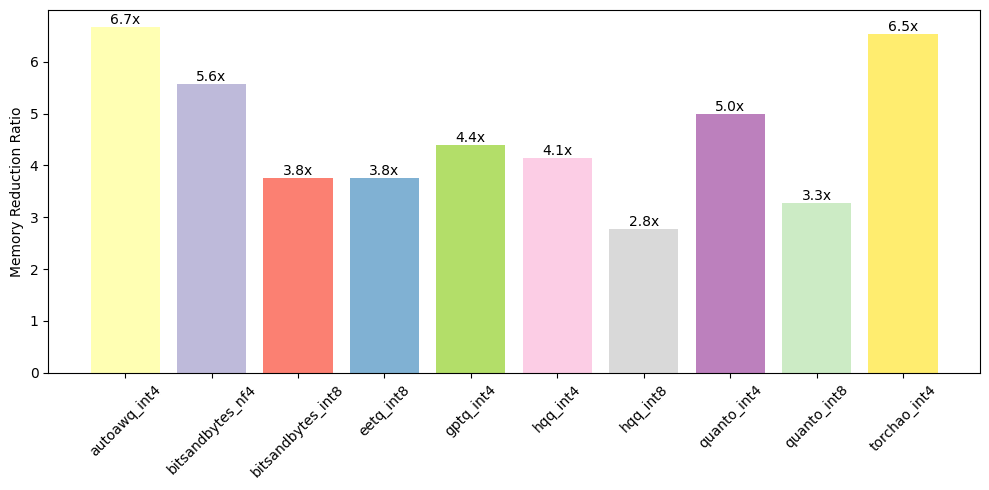}
    \caption{Phi 3.5 Mini Instruct}
    \label{fig:memory_phi}
  \end{subfigure}

  \begin{subfigure}[b]{\linewidth}
    \centering
    \includegraphics[width=0.8\linewidth]{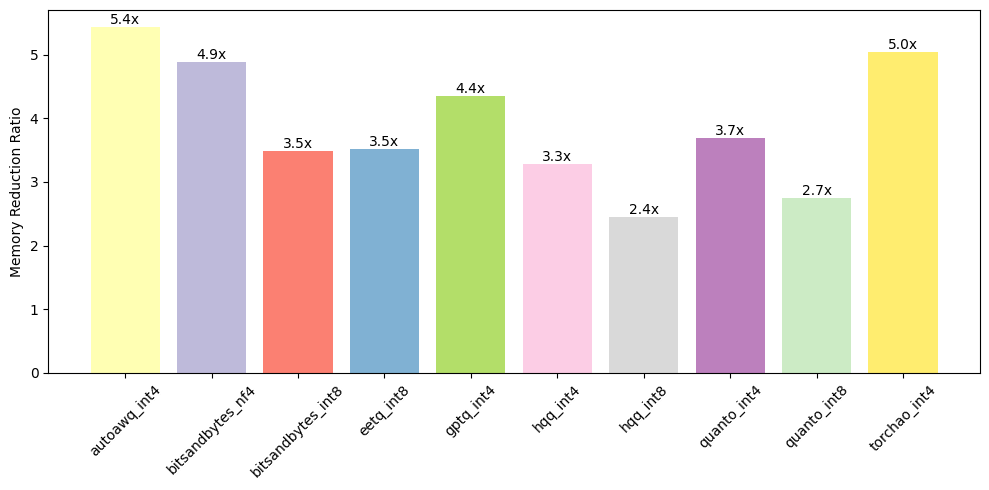}
    \caption{Qwen 2.5 7B}
    \label{fig:memory_qwen}
  \end{subfigure}

  \caption{Memory Efficiency Ratio Across Models}
  \label{fig:memory_graphs}
\end{figure}

While Table~\ref{tab:memory_efficiency_ratio} provides a concise summary of average memory reduction ratios, Figure~\ref{fig:memory_graphs} highlights important variations across models. For example, AutoAWQ int4 achieves a 6.7$\times$ memory reduction on Phi~3.5~Mini, but only 5.1$\times$ on Llama 2 7B. Similarly, TorchAO int4 consistently performs well across models, whereas HQQ int4 and Quanto int8 exhibit more fluctuation in efficiency. These differences reflect how quantization interacts with model-specific factors such as weight layout, parameter sparsity, and layer normalization variants. Consequently, practitioners should consider both quantization method and model architecture when targeting memory-constrained environments.

\section{Cost \& Emissions}

\begin{figure}[h!]
    \centering
    \includegraphics[width=0.8\textwidth]{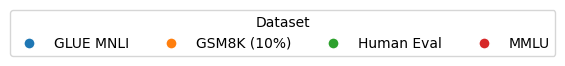}\\[0.5em]
    \includegraphics[width=\textwidth]{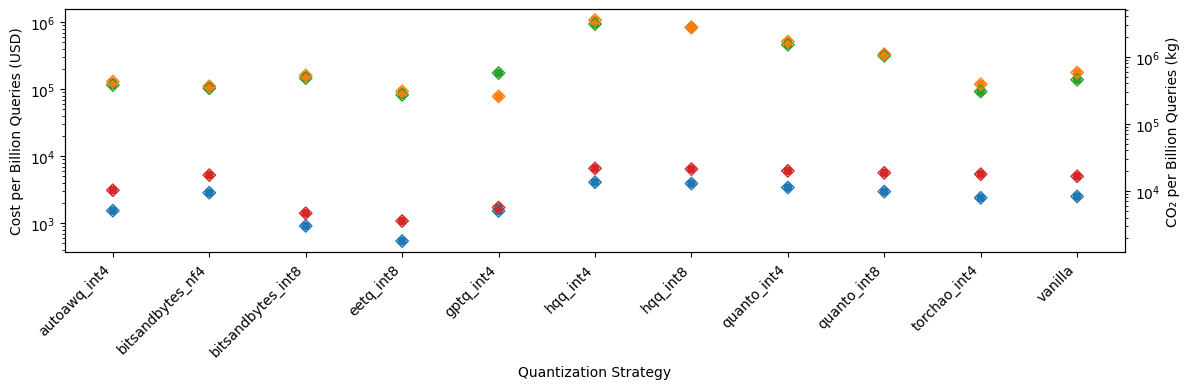}
    \caption{Cost \& CO\textsubscript{2} per Billion Queries by Quantization Strategy and Dataset on A6000 (log scaled)}
    \label{fig:cost_co2}
\end{figure}

Figure~\ref{fig:cost_co2} presents the estimated monetary cost (USD) and carbon emissions (kg~CO\textsubscript{2}) per billion queries for each quantization strategy across all 4 evaluated datasets on an Nvidia A6000 GPU.
The left $y$-axis shows the cost per billion queries, while the right $y$-axis corresponds to the equivalent CO\textsubscript{2} emissions. 
Both axes use a logarithmic scale to improve the visibility of lower-magnitude values.

We assume the workloads are executed in the AWS us-east-1 region (Northern Virginia), using a regional electricity price of \$0.093 USD per kWh \cite{brightlio}, and a carbon emission value of 0.393 kg CO\textsubscript{2} per kWh \cite{epa}.

We observe that the cost and emissions vary significantly between quantization strategies. Bitsandbytes int8 and EETQ int8 yield the lowest costs and emissions across all datasets, whereas certain 4-bit methods such as HQQ int4 can be substantially more expensive in scenarios where their kernel implementations are less optimized for the target hardware. The relative ordering of strategies is consistent across datasets, with differences in magnitude driven largely by per-query runtime and energy consumption.

These results emphasize that quantization choice has a direct and often large impact on both financial and environmental inference costs. Selecting an efficient quantization scheme can reduce CO\textsubscript{2} emissions by orders of magnitude without sacrificing accuracy, as discussed in Section~\ref{sec:accuracy_degradation}.

\chapter{Conclusion}

In this work, we presented an extensive analysis of energy efficiency for quantized LLM inference, evaluating 10 quantization strategies across three models and four diverse tasks. We demonstrated that modern 4-bit and 8-bit weight-only quantization techniques can dramatically reduce joules per token during generation. In particular, fused low-bit methods such as EETQ int8 and Bitsandbytes int8 consistently provided the best energy-accuracy trade-offs, achieving up to $4\times$ energy savings for single-token outputs, with EETQ sustaining nearly $2\times$ savings even for long sequences.

Importantly, int8 quantization was able to preserve model accuracy within 2.5\% of full-precision baselines, showing that efficiency gains no longer require major accuracy sacrifices at 8-bit precision, although 4-bit methods still pose risks. Our findings reinforce that quantization is not only an optimization, but a practical and powerful tool for making LLM services significantly more sustainable.

From a practical engineering perspective, our results indicate that latency serves as a reliable proxy for total energy consumption. This correlation is driven by the high static power use of modern GPUs. Since an idle GPU maintains a substantial baseline power draw, reducing the total runtime of an inference task is often more impactful than minimizing instantaneous power consumption. Consequently, practitioners can effectively optimize for energy efficiency by prioritizing low-latency execution. We further observe that increasing batch sizes can moderately amplify these efficiency gains by enhancing hardware utilization and distributing fixed architectural overheads over a greater volume of processed tokens.

Finally, the most important and novel finding of this study is that quantization strategies exhibit peak energy efficiency relative to full-precision baselines on long input, short output workloads. We provide an intuitive architectural explanation for this phenomenon based on the varying arithmetic intensity of LLM inference phases. During the prefill phase, the large matrix-matrix multiplications required for prompt processing allow the overhead of dequantization to be effectively amortized over a high volume of floating-point operations. In contrast, the auto-regressive decoding phase relies on smaller matrix-vector operations where dequantization costs represent a much larger fraction of the total execution budget. Therefore, as input lengths increase, the system spends a greater proportion of its energy in the computationally dense prefill stage, where the advantages of quantized computations most clearly outweigh their overheads.

\section{Future Work}

Building on the empirical findings of this study, this section outlines several promising directions for future research. The following subsections explore avenues for further optimizing LLM sustainability through intelligent resource management, expanded quantization scopes, and hardware-specific advancements.

\subsection{Context-Aware Scheduling and Precision Allocation}
One promising direction is the development of intelligent scheduling policies that dynamically select model precision based on the requirements of each request. For example, short-answer or less complex queries can be routed to an efficient 8-bit quantized model, while longer or mission-critical queries can be directed to a full-precision model to maximize accuracy. This approach leverages the observation that quantized models often retain high accuracy while consuming significantly less energy. A key insight is that energy cost scales with output length, since each generated token consumes a roughly constant amount of energy for a given model. Therefore, if a request is expected to produce a lengthy output (e.g., a narrative or code completion), a scheduler could assign it to a full-precision model to maximize accuracy. Conversely, tasks with low token output such as multiple-choice questions can be assigned to a quantized model, reducing energy use without compromising quality. This adaptive allocation reduces energy consumption while maintaining output quality by assigning the appropriate model to each task.  

In addition to precision selection, cloud providers can incorporate carbon-aware orchestration. Many data centers now monitor the carbon intensity of their power supply throughout the day. Non-urgent or batch inference jobs can be deferred to times when the grid is cleaner, or migrated to regions where renewable energy is more abundant \cite{carbon_aware}. By shifting flexible workloads to off-peak hours or greener geographic locations, schedulers can lower emissions and energy costs with minimal impact on latency. In summary, intelligent scheduling that combines precision allocation with carbon-aware workload placement can enable large cloud data centers to serve LLM requests more sustainably.

\subsection{Extending Quantization Beyond Model Weights}
Another direction is to extend quantization beyond model weights to activations and the KV cache, thereby unlocking greater efficiency. Most current LLM quantization focuses on compressing model weights to lower bit-widths. However, for long-context or streaming inference, the memory overhead from activations, particularly the KV cache in Transformers, becomes a bottleneck. For instance, storing a 10,000-token context for a 7B model in FP32 can require $\sim$10 GB of memory, whereas an 8-bit cache would reduce this to $\sim$2.5 GB. By compressing the cache into 8-bit or even 4-bit formats, LLMs can support much longer inputs and outputs on the same hardware budget.

Beyond the cache, quantizing layer activations throughout the model is a compelling but challenging step. A fully quantized deployment, where most matrix multiplications use int8 or int4 arithmetic end-to-end, would enable faster inference and lower energy consumption per token, allowing more requests to be served on the same hardware. The main challenge is preserving model fidelity, as aggressive quantization of activations can introduce numerical error or instability due to outlier values in intermediate layers. Nevertheless, extending quantization beyond weights is a promising avenue for further reducing the runtime costs of LLM inference.

\subsection{Low-Bit Hardware and Accelerator Optimization}
On the hardware side, future work could evaluate quantized inference on TPUs \cite{tpu} and emerging AI accelerators optimized for low-bit operations to examine energy trends. Such directions offer practical pathways to extend and apply these findings, with the goal of advancing scalable and sustainable LLM deployment.

\subsection{Sources of Energy Efficiency in Quantized Inference}
A deeper understanding of the specific origins of energy savings in quantized inference is essential for guiding the design of more efficient models, quantization techniques, and hardware. Future work should explicitly separate the effects of reduced memory-access cost from the effects of lower-precision computation. Quantization decreases the number of bytes transferred through the memory hierarchy and also changes the arithmetic performed on tensor cores, but the relative contribution of each factor to overall energy reduction is not yet well understood.

A systematic breakdown of memory traffic, compute activity, and power consumption during inference would help clarify where the dominant energy savings arise. Such insights would guide the development of quantization methods that target the most influential components of the energy budget and inform future designs that better align with the actual sources of efficiency gains.

\subsection{Quantization Configuration}
While our study provides valuable insights into the energy efficiency and accuracy trade-offs of various quantization strategies, it does not fully explore the role of quantization granularity or symmetry. Future research could investigate the impact of different granularity or symmetry settings within each quantization strategy, as the optimal configuration may be task-dependent and interact with other hyperparameters. A more thorough understanding of these parameters could lead to further improvements in quantized LLM performance.

\subsection{Overall Outlook}
In summary, the four main directions — smarter scheduling, broader quantization, hardware specialization, and quantization configuration — converge toward scalable and sustainable deployment of large Transformer-based models. Intelligent workload routing and precision selection allow cloud data centers to reduce waste while maintaining service quality. Extending quantization beyond weights lowers the resource footprint of each inference with minimal accuracy loss, and aligning these methods with low-bit hardware ensures energy savings translate into real-world impact. As inference continues to dominate machine learning energy consumption, advances in these areas are both timely and significant. Together, they provide a roadmap for deploying large language models responsibly, balancing efficiency with the quality users expect.

\bibliographystyle{plain}
\cleardoublepage %
\phantomsection  %
\renewcommand*{\bibname}{References}

\addcontentsline{toc}{chapter}{\textbf{References}}

\bibliography{bibliography.bib}

\end{document}